%% file: main.tex
\documentclass[a4paper,fleqn]{cas-sc}

\usepackage[numbers]{natbib}

\def\tsc#1{\csdef{#1}{\textsc{\lowercase{#1}}\xspace}}
\tsc{WGM}
\tsc{QE}
\tsc{EP}
\tsc{PMS}
\tsc{BEC}
\tsc{DE}
\usepackage{xspace}
\usepackage{pifont}
\usepackage{longtable}
\usepackage{tabularray}
\usepackage[dvipsnames]{xcolor}
\usepackage{tikz}
\usetikzlibrary{shapes.geometric, arrows}

\tikzstyle{startstop} = [rectangle, rounded corners, minimum width=3cm, minimum height=1cm,text centered, draw=black, fill=red!30]
\tikzstyle{process} = [rectangle, minimum width=3cm, minimum height=1cm, text centered, draw=black, fill=orange!30]
\tikzstyle{decision} = [diamond, minimum width=3cm, minimum height=1cm, text centered, draw=black, fill=green!30]
\tikzstyle{arrow} = [thick,->,>=stealth]
\usepackage{graphicx}
\graphicspath{{figs/}{figs/convergence/}{figs/trajectories/}{figs/wins}}

\definecolor{graytblr}{RGB}{240,240,240}

\newcommand{\agsk}{\texttt{AGSK}}
\newcommand{\agskimode}{\texttt{APGSK}}
\newcommand{\birmin}{\texttt{BIRMIN}}
\newcommand{\de}{\texttt{DE}}
\newcommand{\eaeig}{\texttt{EA4eig}}
\newcommand{\ebocmar}{\texttt{EBOwCMAR}}
\newcommand{\elshade}{\texttt{ELSHADE}}
\newcommand{\dtcg}{\texttt{I-DTC-GL}}
\newcommand{\lshade}{\texttt{LSHADE}}
\newcommand{\nm}{\texttt{NM}}
\newcommand{\pso}{\texttt{PSO}}
\newcommand{\spso}{\texttt{SPSO}}
\newcommand{\start}{\texttt{start}} %if we want to change later
\newcommand{\goal}{\texttt{goal}}

\newcommand{\directgolib}{\texttt{DIRECTGOLib}}

\begin{document}
\let\WriteBookmarks\relax
\def\floatpagepagefraction{1}
\def\textpagefraction{.001}

\shorttitle{Benchmarking global optimization techniques for UAV path planning}

\shortauthors{Shehadeh and Kudela}

\title[mode=title]{Benchmarking global optimization techniques for unmanned aerial vehicle path planning}
%\tnotemark[1,2]

%\tnotetext[1]{This document is the results of the research project funded by the National Science Foundation.}

%\tnotetext[2]{The second title footnote which is a longer text matter to fill through the whole text width and overflow into another line in the footnotes area of the first page.}


\author[1]{Mhd Ali Shehadeh}[orcid=0000-0001-5550-4650]

\credit{Methodology, Software, Formal analysis, Validation, Writing--Original draft, Visualization}
\author[2]{Jakub Kůdela}[orcid=0000-0002-4372-2105]
\cormark[1]
\ead{Jakub.Kudela@vutbr.cz}
\credit{Conceptualization, Methodology, Software, Formal analysis, Validation, Writing--Original draft, Visualization, Funding acquisition, Project administration, Supervision}

\affiliation[1]{organization={Institute of Automation and Computer Science, Brno University of Technology},
                addressline={Technick\'{a} 2869/2},
                city={Brno},
                postcode={61669}, 
                country={Czech Republic}
               }
\affiliation[2]{organization={Institute of Automation and Computer Science, Brno University of Technology},
                addressline={Technick\'{a} 2869/2},
                city={Brno},
                postcode={61669}, 
                country={Czech Republic}
               }

\cortext[cor1]{Corresponding author}
%\cortext[cor2]{Principal corresponding author}

\begin{abstract}
The Unmanned Aerial Vehicle (UAV) path planning problem is a complex optimization problem in the field of robotics. In this paper, we investigate the possible utilization of this problem in benchmarking global optimization methods. We devise a problem instance generator and pick 56 representative instances, which we compare to established benchmarking suits through Exploratory Landscape Analysis to show their uniqueness. For the computational comparison, we select twelve well-performing global optimization techniques from both subfields of stochastic algorithms (evolutionary computation methods) and deterministic algorithms (Dividing RECTangles, or DIRECT-type methods). The experiments were conducted in settings with varying dimensionality and computational budgets. The results were analyzed through several criteria (number of best-found solutions, mean relative error, Friedman ranks) and utilized established statistical tests. The best-ranking methods for the UAV problems were almost universally the top-performing evolutionary techniques from recent competitions on numerical optimization at the Institute of Electrical and Electronics Engineers Congress on Evolutionary Computation. Lastly, we discussed the variable dimension characteristics of the studied UAV problems that remain still largely under-investigated.

\end{abstract}

% \includegraphics{figs/grabs.pdf}

% \begin{highlights}
% \item Novel formulation of conceptual ship design.
% \item Comparison of state-of-the-art and canonical metaheuristics.
% \item The best methods for the problem were LSHADE and DE.
% \end{highlights}

% Keywords

\maketitle

\input{Introduction.tex}

\input{Problem_Description.tex}

\input{Results.tex}

\input{Conclusion.tex}
\input{Acknowledgement.tex}

\appendix
\printcredits

%\bibliographystyle{elsarticle-num}
\bibliographystyle{model1-num-names}

\nocite* % adds all citations in the bib file (just to make sure it works)
\bibliography{main}

\end{document}

%% file: Introduction.tex
\section{Introduction}
Over the last few years, we have seen a formidable improvement in the utilization and availability of Unmanned Aerial Vehicles (UAVs). UAVs have become more prevalent in diverse fields such as tracking \cite{kamate2015application}, meteorology \cite{sziroczak2022review}, monitoring \cite{giordan2017use}, search and rescue \cite{yang2020maritime}, security and surveillance \cite{ibrahim2010moving},  agriculture \cite{kim2019unmanned}, or in military applications \cite{xiaoning2020analysis}. The majority of tasks exercised by UAVs require some level of autonomy in making decisions based on real-time changes. Among such decisions is also the trajectory selection or path planning \cite{ait2022novel}. For UAVs, successful path planning is critical to their ability to complete required tasks and avoid various threats that might appear in the environment \cite{phung2021safety}. The planned path should also be optimized with respect to a specific criterion defined by the application. For instance, in areas such as surface inspection or aerial photography, the optimization criterion is typically the minimization of the path length between pre-selected locations that the UAV should visit, which also roughly corresponds to the minimization of flight time and required fuel \cite{phung2019system}. Other criteria might involve minimizing the flight time itself \cite{lin2009uav}, maximizing the detection probability \cite{phung2020motion}, or searching for Pareto solutions in multi-objective navigation \cite{yin2017offline}. During the completion of the task, UAVs generally encounter various static or dynamic obstacles, which can lead to physical damage. Avoiding such obstacles and threats is one of the feasibility requirements in AUV path planning. Other feasibility requirements might include alignment of the path with the limitation of the particular UAV, e.g. fuel consumption, flight time, flight altitude, climbing angle, or turning rate.

Various approaches have been proposed for UAV path planning. One of them is graph search, which divides the environment into connected discrete regions. Each of these regions forms a vertex of a graph in which the path is being searched. Voronoi diagrams were used in \cite{beard2002coordinated} for the construction of the graph and the $k$-best path algorithm \cite{eppstein1998finding} was then used for the path planning. A different graph-based approach is the probabilistic roadmap (PRM) which uses sampling to generate the vertices of the graph \cite{pettersson2006probabilistic}. The rapid-exploring random trees algorithm also uses a similar strategy to PRM to create the search graph \cite{lin2017sampling}. Even though these graph-based approaches are very effective in generating feasible paths, they are well-suited when the inclusion of constraints related to various UAV maneuvers is necessary. A different set of approaches based on cell decomposition can also be used. Here the problem is represented as a grid of equal cells and various heuristic search methods are utilized to find the path. One of the popular search algorithms in this context is A* \cite{penin2018minimum}, which can be also extended to handle constraints specific to the UAV \cite{szczerba2000robust}. Cell decomposition-based methods were also used for path coordination between Unmanned Ground Vehicles and UAVs \cite{li2016hybrid}, autonomous flight control \cite{kwak2018autonomous}, or for path predictions in real-time multi-UAV mission planning \cite{sun2015triple}. The major drawback of cell decomposition-based techniques lies in their scalability. The number of cells increases exponentially with the dimension of the search space. Another category of approaches for AUV path planning consists of the potential field methods, which search the continuous space directly (instead of relying on its discretization like the graph and cell decomposition-based methods). In this framework, the UAV is treated as a particle that moves under the influence of an artificial potential field which is constructed based on the objective and the threads \cite{tang2019optimized}. 

In contrast to the abovementioned methods, global optimization approaches have become more prevalent in path planning because of their ability to deal with various UAV constraints and the capacity to search for the global optimum in complex scenarios. The two main branches of global optimization methods are deterministic and stochastic techniques. Of the deterministic global optimization techniques, Lipschitz optimization methods \cite{pinter1995global} are among the most classical ones. The most well-known and widely utilized extension of the Lipschitz methods is the DIRECT (DIviding RECTangles) algorithm \cite{jones1993lipschitzian}, which was found to be effective on low-dimensional real-world problems. However, the major drawback of DIRECT seemed to be its lack of exploitation capabilities \cite{sergeyev2006global}. Among the stochastic global optimization techniques (also called metaheuristics) currently the most popular ones are evolutionary computation (EC) algorithms (also called nature-inspired or bio-inspired algorithms). These methods imitate biological processes such as natural selection, or evolution, where solutions are represented as individuals that reproduce and mutate to generate new, potentially improved candidate solutions for the given problem \cite{molina2020comprehensive}. Other EC methods try to mimic the collective behavior of simple agents, giving rise to the concept of swarm intelligence \cite{yang2013swarm}. However, many of the newly proposed methods have been found to be a “re-branding” of older methods \cite{camacho2020grey,camacho2023exposing}, have dubious quality \cite{aranha2022metaphor,campelo2023lessons}, or used biased experiments for the analysis of their performance \cite{kudela2022commentary, kudela2022critical,kudela2023evolutionary}.

Recently, we have seen increased interest in using various EC methods for UAV path planning, with earlier works utilizing ``standard'' EC techniques such as particle swarm optimization (\pso) \cite{roberge2012comparison}, artificial bee colony \cite{xu2010chaotic}, differential evolution (\de) \cite{fu2013route}, or ant colony optimization \cite{yu2018aco}. For our work, the main point of reference is the paper \cite{phung2021safety}, where the spherical vector-based \pso{} (\spso) for safety-enhanced UAV path planning was proposed. In the paper, the authors devised the framework for the UAV path planning problem, including an objective function that combined path length, obstacle avoidance, and altitude and smoothness penalties. There are several papers that utilized the model proposed in \cite{phung2021safety}. Many of them expand the \spso{} algorithm, for instance, by adding a chaotic generator as in \cite{chu2022chaos}, or a multi-subgroup Gaussian mutation operator and elite individual genetic strategy as in \cite{wang2023sggtso}. Another set of work investigated hybridization strategies of \pso{} with differential evolution \cite{huang2023adaptive} or simulated annealing \cite{yu2022novel}. Some papers use different variants of the notorious nature-inspired methods, such as Grey wolf optimizer, Beluga whale optimizer, and similar ones, which however (as discussed in the previous paragraph), should not be considered in the first place. All of the papers rely on the same problem instances that were proposed in \cite{phung2021safety}.

As many global optimization methods are hard to analyze analytically, their utility is usually analyzed through benchmarking \cite{hellwig2019benchmarking}. Although various benchmark sets and functions were introduced \cite{garcia2017since,kudela2022new}, the most widely used benchmark sets have been developed for special sessions (competitions) on black-box optimization in two EC conferences: the IEEE Congress on Evolutionary Computation (CEC) and the Genetic and Evolutionary Conference (GECCO), where the Black-Box Optimization Benchmarking (BBOB) workshop was held \cite{hansen2021coco}. The utilization of these benchmark suits is not without criticism, mainly concerning the artificial nature of the problems \cite{piotrowski2015regarding}, and recommended testing these methods on real-world instances instead \cite{tzanetos2021nature}. Expanding the existing benchmark suits with new problems will enhance algorithmic comparisons \cite{piotrowski2025metaheuristics}. One possible method for comparing the characteristics of benchmark suits is the utilization of the Exploratory Landscape Analysis (ELA) \cite{mersmann2010benchmarking}. In this method, the various benchmark functions are described by a set of numerical landscape features, which represent different aspects of these functions. 

\begin{figure}[!ht]
    \centering
    \includegraphics[width=0.9\linewidth]{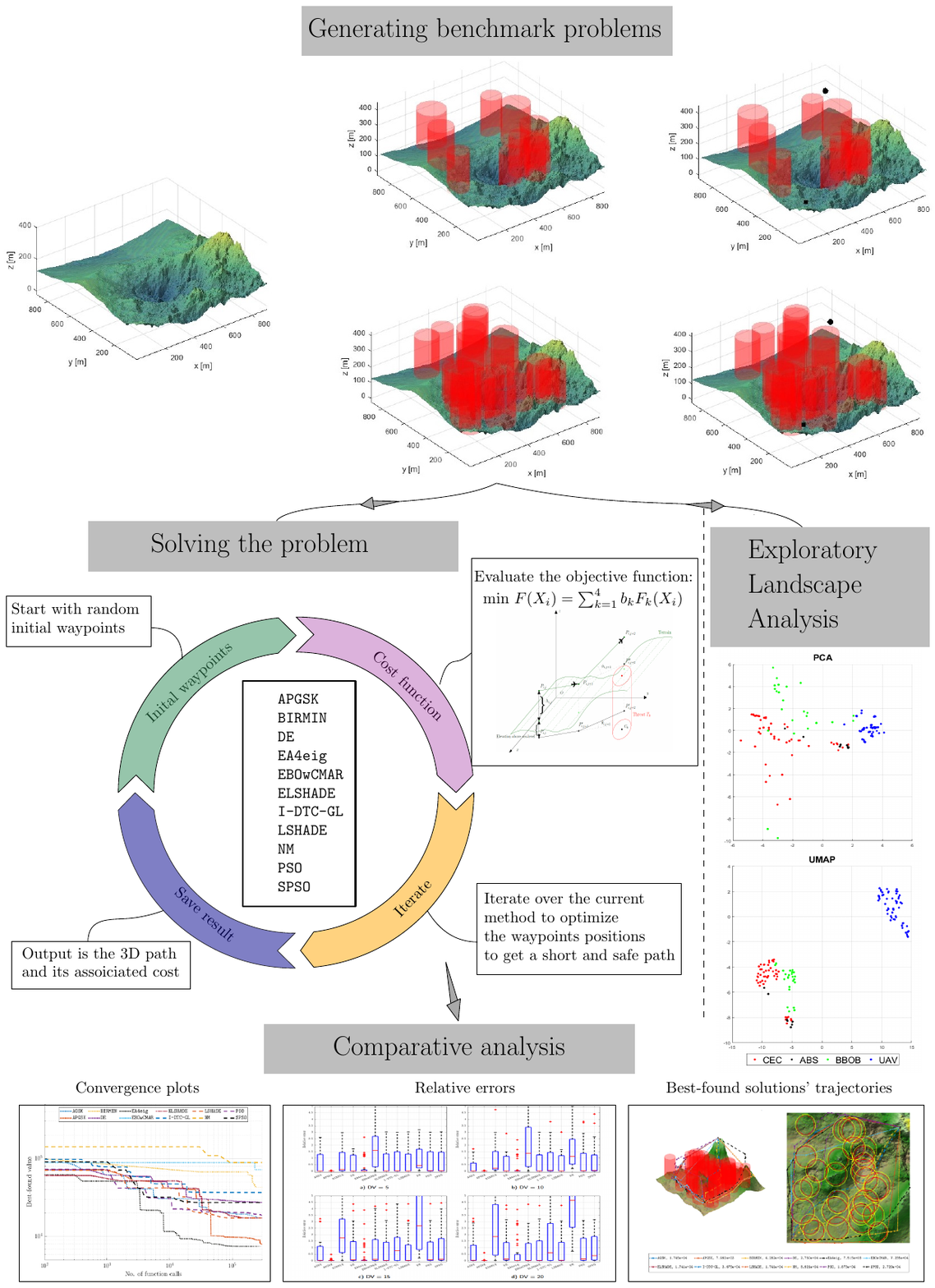}
    \caption{General scheme of the adopted approach.}
    \label{fig:general}
\end{figure}

For the majority of optimization problems, the dimension of the problem is considered fixed. However, there are situations, in which the appropriate dimensionality of the problem is not known in advance \cite{kudelacec}. Various real-world applications, where approaches such as Messy genetic algorithms \cite{goldberg1989messy}, which utilize a variable-length encoding, have been found to outperform their fixed dimension counterparts. Recent applications of such methods include scheduling pump operations \cite{karami2020using}, or truss topology and geometry optimization \cite{khetan2015managing}. The studied UAV problem can be seen to have this variable dimension characteristic.

In this paper, we aim to fill several research gaps both in the use of global optimization methods for UAV path planning and in available benchmarking sets for global optimization methods:
\begin{itemize}
    \item We develop an instance generation procedure and hand-picked 56 instances for numerical comparison.
    \item We perform an ELA features-based comparison of the generated instances with established benchmark suits, showing the novelty of the generated instances.
    \item We conduct an extensive computational comparison of state-of-the-art deterministic and stochastic black-box optimization methods on different computational budgets, discussing the strengths and weaknesses of different techniques in different setups.
    \item We investigate the variable dimension characteristic of the problem, which has so far been seriously understudied.
    \item The code for the instance generation and for running the experiments, and the resulting data, are publicly available at a Zenodo repository \cite{zenodolink}, encouraging the adoption of the generated UAV problem instances for further algorithmic comparisons.
\end{itemize}

The rest of the paper is structured as follows. In Section \ref{sec2}, we review the model proposed in \cite{phung2021safety}, propose a new terrain/instance generation scheme, and provide an ELA-based comparison of the generated instances with established benchmark suits. In Section \ref{sec3}, we briefly describe the global optimization methods used for the numerical comparison and the experimental setup. Results of the experiments are reported and discussed in Section \ref{sec4}. Finally, we conclude with our findings and propose several future research directions in Section \ref{sec5}. The general scheme of the approach adopted in this paper is visualized in Fig.~\ref{fig:general}.

%% file: Problem_Description.tex
\section{Problem Context}\label{sec2}

% 8-4-2024 This investigation centers on the assessment of a model introduced by Phung and Ha, detailed in \cite{spso2021}. Our scrutiny involves subjecting a chosen set of algorithms to testing using a model inspired by their work, specifically designed for UAV path planning. The model formulates a comprehensive cost function that incorporates four pivotal criteria: path optimality, safety, feasibility constraints, maintenance of a specific average height above the terrain, and the smoothness of the planned path.

%To ensure a robust evaluation environment for these diverse algorithms, a varied set of terrains featuring different threats is imperative. Consequently, we developed a terrain generation tool capable of randomly creating terrain and obstacles. From the pool of 56 instances generated, each exhibiting distinct topological characteristics such as plains, hills, steep inclines, and deep holes, we carefully selected instances that present intriguing challenges for thorough examination.
Consider a UAV navigating from an initial to a destination point in a known environment with obstacles and threats such as mountains, buildings, radar installations, and ground-to-air missiles (that may jeopardize the UAV's motion or completely destroy it if it passes/collides with the threat). This scenario presents a classical path planning problem. The objective here is to find the shortest path while safely maneuvering the terrain.

% This problem can be solved by discretizing the path into a series of waypoints. The UAV then traverses the path that is linearly connecting these waypoints. Thus, the problem can be formulated as a discrete constrained optimization problem. The optimization process in this paper aims to minimize a cost function that incorporates various safety criteria which are introduced by Phung and Ha \cite{spso2021}. 

% While this investigation builds on the assessment of a model introduced by Phung and Ha, detailed in \cite{spso2021}. Our research involves subjecting a chosen set of state-of-the-art algorithms, which are known to perform better in various optimization problems, to testing using a model specifically designed for UAV path planning. 

The optimization model, adopted from \cite{spso2021}, formulates a comprehensive cost function that incorporates four pivotal criteria: path optimality, safety, feasibility constraints, maintenance of a specific average height above the terrain, and the smoothness of the planned path.

As mentioned earlier, our path planning problem is set for already-known terrains. To ensure a robust evaluation environment for a diverse set of algorithms, a diverse set of terrains featuring different threats is imperative. We employed a terrain generation tool capable of randomly creating terrain and obstacles. Then, from a pool of 5000 generated instances, each exhibiting distinct topological characteristics such as plains, hills, steep inclines, and deep valleys, we carefully selected 56 instances that present intriguing challenges for a thorough numerical examination.

\subsection{Terrain Generation}
\label{s56}
The terrain generation function presented in this paper utilizes a tool \cite{matlab-terrain} for generating approximately realistic-looking terrains. A set of points approximating terrain features is plotted given an input set of parameters, which are characterized by its elevation, roughness, and spatial distribution. The function's inputs include the number of iterations (\texttt{$n$}), the size of the output mesh (we set it to be nine hundred pixels in all terrains, which is the same size as in \cite{phung2021safety}), initial elevation (\texttt{$h_0$}), initial roughness (\texttt{$r_0$}), and varying roughness through the terrain (\texttt{$rr$}).

% The key steps of the algorithm are as follows:

% \begin{itemize}
%   \item Generate initial coordinates and roughness values for a set of points.
%   \item Iteratively create new points based on the characteristics of existing points.
%   \item Normalize the distribution of points about the median.
%   \item Optionally, create a mesh over the terrain for visualization.
% \end{itemize}
Changing these input parameters or the randomness seed results in a large variety of generated artificial terrains. We ran this algorithm with five thousand different combinations of parameters and hand-picked twenty-eight terrains that displayed interesting and ``real-life looking'' characteristics.

\begin{figure}
    \centering
    \begin{tabular}{cc}
       \includegraphics[width = 0.45\linewidth]{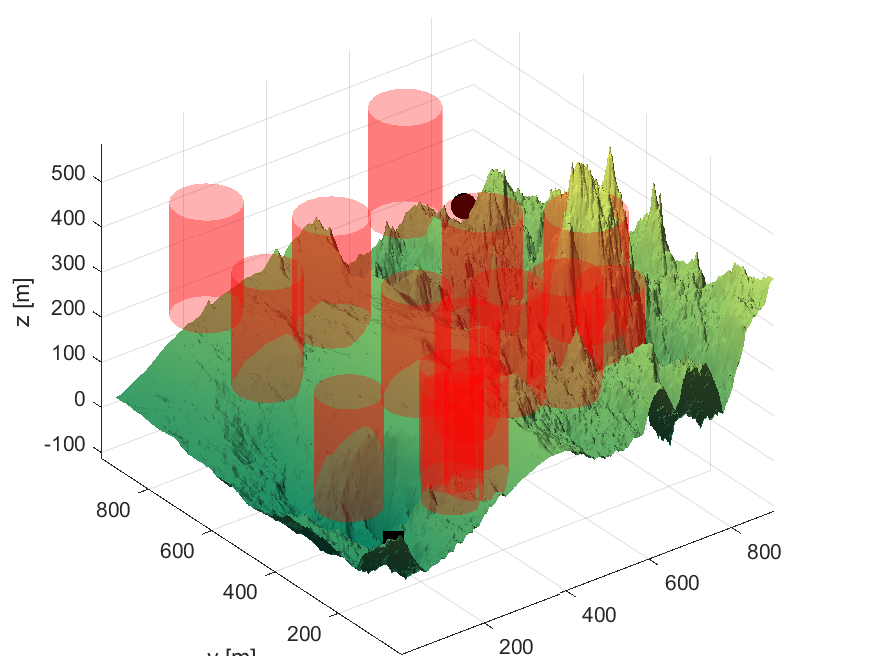}  &  \includegraphics[width = 0.45\linewidth]{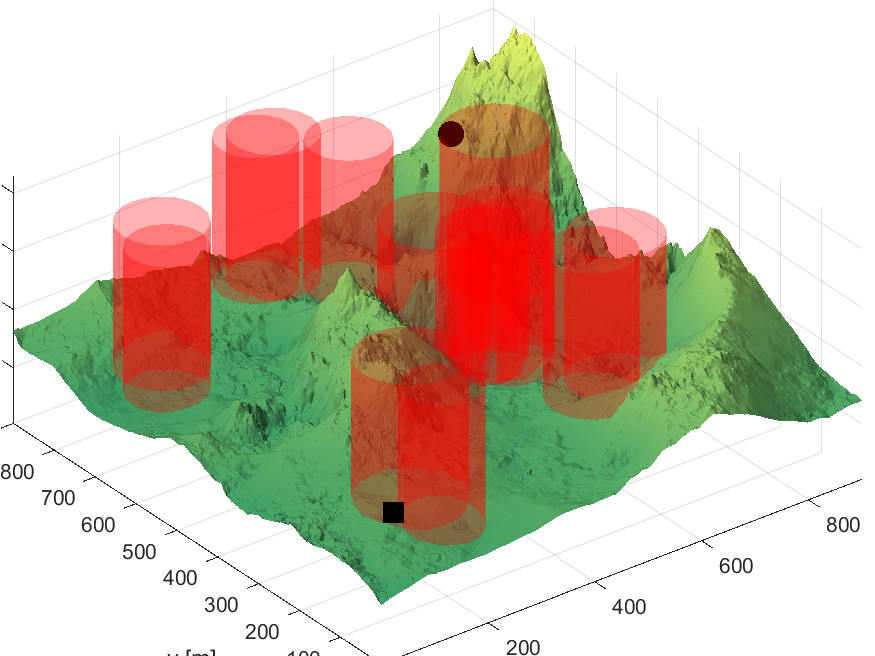}\\
       a) $r$=-60, $rr$=300, seed=13, num=15 & b)  $r$=-20, $rr$=300, seed=3, num=15\\[3mm]
       \includegraphics[width = 0.45\linewidth]{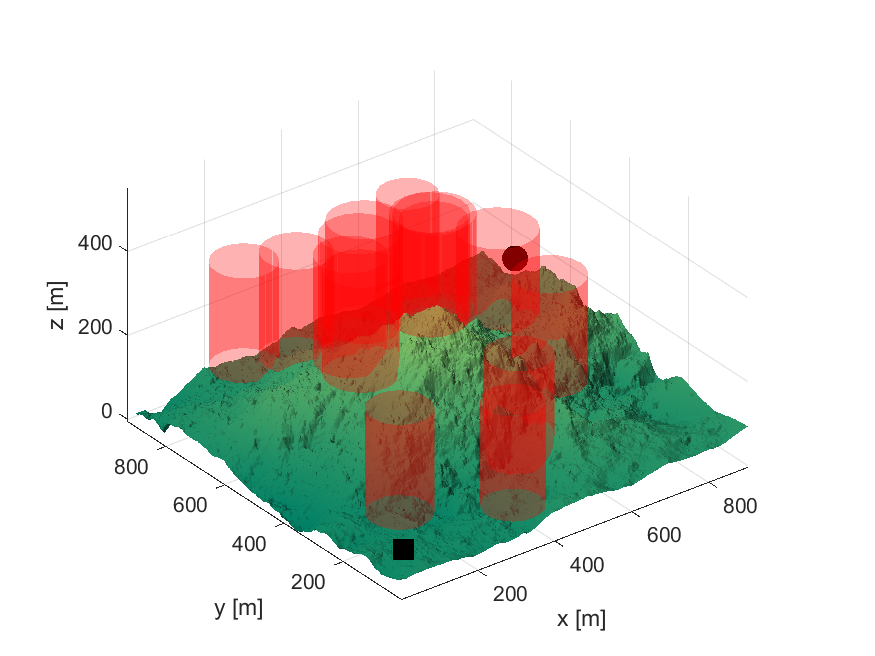} & \includegraphics[width = 0.45\linewidth]{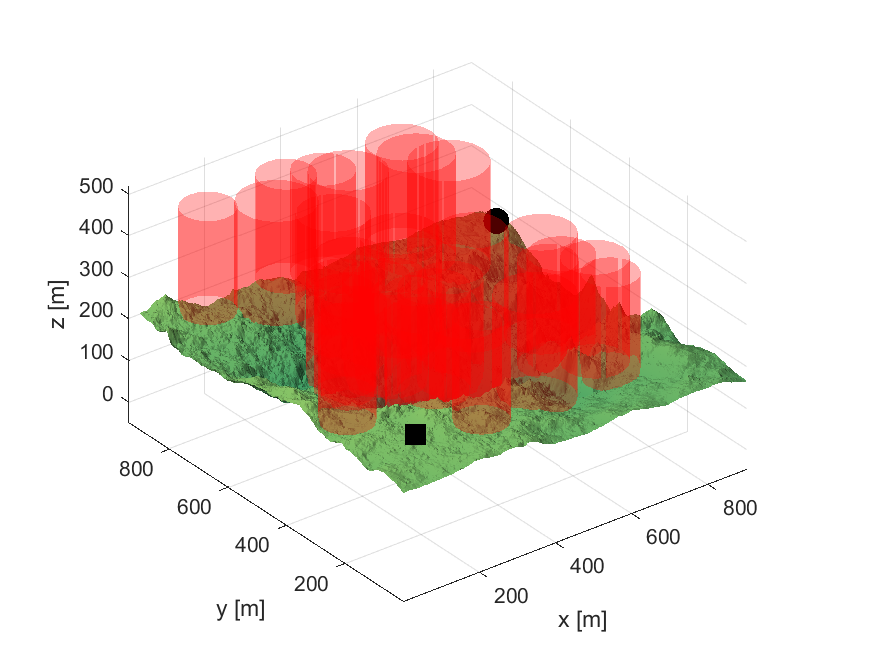}\\
       c)  $r$=-60, $rr$=60, seed=18, num=15 & d)  $r$=60, $rr$=-60, seed=18, num=30\\
    \end{tabular}
    \caption{A Side view of four generated terrains with 900 pixels. The black rectangle/circle resembles the \start{}/\goal{} points respectively. The red cylinders represent the threats that the UAV must avoid.}
    \label{fig:choosen}
\end{figure}

%The chosen terrain instances are particularly useful for testing algorithms related to path planning for Unmanned Aerial Vehicles (UAVs) in diverse environments. The chosen set encompasses variations in elevation, roughness, and topological features such as plains, hills, steep inclines, and deep holes, look at the exemplary photos attached in Fig. \ref{fig:choosen}.
The chosen terrain instances serve as benchmark problems, which are essential for evaluating the performance of different methods in the field of path planning for UAVs across diverse environments. Exemplary terrains that were generated by the procedure can be seen in Fig. \ref{fig:choosen}.

After selecting the set of terrains, we randomly generate cylindrical threats with different heights and radii and then place the \start{} and \goal{} points to be outside these threats and in opposite corners of the artificially generated terrain. In terms of obstacle complexity, two different variations of threats for each terrain were considered. The first one with fifteen threats and the denser second one with thirty threats. This results in a total of fifty-six instances for comparison.
%in oreder to motion plan the UAV from the start to the end point we simplify its journey by discreitizing its path into a number of waypoints. and then we evaluate the otimilatuty of these points by a set of methods. The core problem of our study is how to navigate the UAV robot through the above-mentioned diverse terrains safely and within the shortest possible path.
% These fifty-six benchmarks problems are then evaluated by the chosen set of global optimization algorithms. These methods are compared at the same number of decision variables $DV = {5, 10, 15, 20}$. Each method employs the same objective function and identical parameters, including population size (fixed at 500),Penalty of passing a threat (set at 10,000), and a maximum number of evaluations (proportional to the number of decision variables), ensuring comparability of the results.

\subsection{Objective Function Formulation}
%The objective function is an import phase for the UAV path planning process, which is the second step after space modeling.
The objective function can be defined as a mathematical formula accommodating specific requirements that need to be optimized. In our case, we adapt the objective function proposed in \cite{Cheng2021} which encompasses four criteria to ensure safety and path optimality. %(del: which are described as follows.) %here we adapt the same objective function and reintroduce it briefly. since our main focus is to test a chosen set of methods over $n \in \{5,10,15,20\}$ Decision Variables.

\subsubsection{Path Length Cost}
The operational efficiency of UAVs relies on the judicious choice of the optimality criterion of the paths adapted to specific applications. Focusing on tasks such as aerial photography, mapping, and surface inspection, minimizing traveled distance yields less time and fuel consumption. Since the path consists of linear segments connecting the waypoints, the criterion for an optimal path, with the shortest geometric distance, is simply to minimize the Euclidean distance between these waypoints.

Let the flight path $X_i$, of an individual solution $i$ be represented as a sequence of $n$ waypoints, each waypoint corresponds to a path node in the search map with coordinates $P_{ij} = (x_{ij}, y_{ij}, z_{ij})$. 

%The UAV's journey is intricately computed to reduce the Euclidean distance between consecutive waypoints and the source and goal points as well.

The cost function $F_1(X_i)$ penalizing the path length between consecutive waypoints (plus the \start{} $S$ and \goal{} $G$ points as well) is expressed as:
\begin{equation}
 F_1(X_i) =\lVert \overrightarrow{P_{iS}P_{i1}} \rVert+ \sum_{j=1}^{n-1} \lVert \overrightarrow{P_{ij}P_{i,j+1}} \rVert +\lVert \overrightarrow{P_{in}P_{iG}} \rVert.
\end{equation}
%, the optimal path criteria is simply chosen to minimise the length of each particle's flight path ($i$ or $X_i$) in the swarm. every flight path $X_i$ is represented as a sequence of $n$ waypoints, each waypoint corresponds to a path node in the search map with coordinates $P_{ij} = (x_{ij}, y_{ij}, z_{ij})$. 

%The UAV's journey is intricately computed to reduce the Euclidean distance between consecutive waypoints and the source and goal points as well.

%The cost function $F_1(X_i)$ governing path length is expressed as:
%\[ F_1(X_i) =\lVert \overrightarrow{P_{is}P_{i,1}} \rVert+ \sum_{j=1}^{n-1} \lVert \overrightarrow{P_{ij}P_{i,j+1}} \rVert +\lVert \overrightarrow{P_{in}P_{i,g}} \rVert \]
\subsubsection{Obstacle Avoidance Cost}

Beyond path-length optimality, the planned path must prioritize the safety of UAV operations. This involves deftly maneuvering around potential threats within the operational space. To ensure that the UAV is passing safely from \start{} to \goal{} destinations without intersecting any threat, we examine each line segment between consecutive waypoints and penalize it if it passes through a threat or approaches its surrounding dangerous zone. An associated penalty $T_k$ is calculated for each threat based on its distance from the line segment $d_k$. 

\begin{figure}
    \centering
    \includegraphics[width=0.5\linewidth]{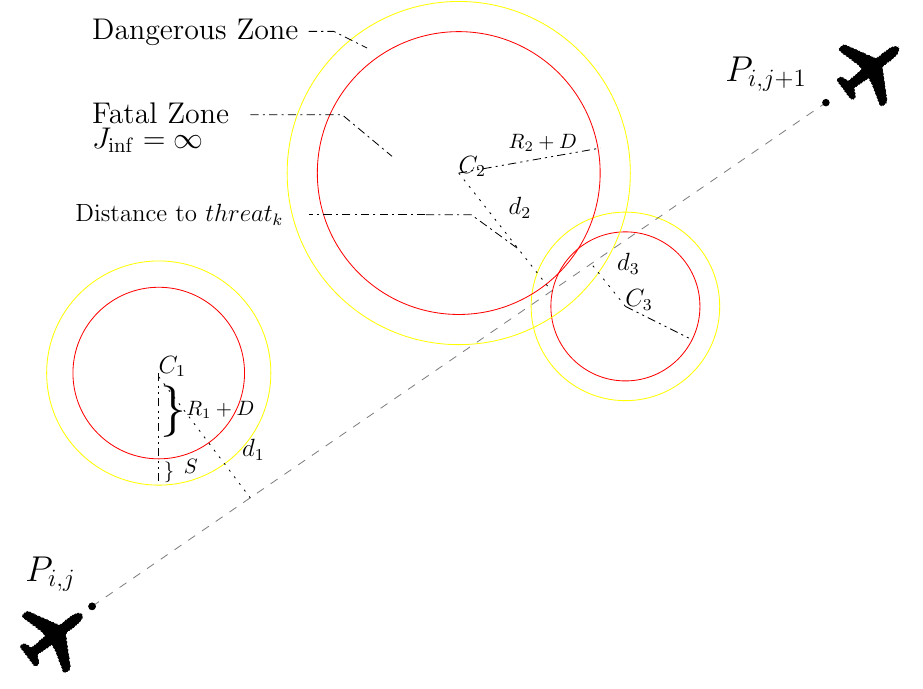}
    \caption{Different scenarios considered in the penalty function $T_k$ for UAV passing a line segment from point $P_{ij}$ to $P_{i,j+1}$. The red circle represents the area of collision with penalty $J_{\text{pen}}$, while the yellow circle indicates the surrounding dangerous zone.}
\label{fig:threat_cost}.
\end{figure}

Threats are modeled as cylindrical entities, with their projection defined by center coordinates $C_k$ and radii $R_k$ as shown in Fig.\ref{fig:threat_cost}. The penalty function $T_k(\overrightarrow{P_{ij}P_{i,j+1}})$ considers the threat distance ($d_k$), UAV diameter ($D$) which is added to each threat radius ($R_k$), and an extended distance ($S$) to represent the dangerous zone:
\begin{equation}
T_k\left(\overrightarrow{P_{i j} P_{i, j+1}}\right)= \begin{cases}0, & \text { if } d_k>S+D+R_k \\ \left(S+D+R_k\right)-d_k, & \text { if } D+R_k<d_k \leq S+D+R_k \\ J_{\text{pen}}, & \text { if } d_k \leq D+R_k.\end{cases}.
\end{equation}
In the exemplary Fig. \ref{fig:threat_cost}, each scenario of $T_k$ is depicted: passing away from the threat (left), passing through the danger zone (middle), and ultimately colliding (right).

The associated penalty for each line segment is the sum of all threat penalties in the threat set $K$. The obstacle avoidance cost $F_2$ adds the cost of all line segments $\overrightarrow{P_{ij}P_{i,j+1}}$ as follows:
\begin{equation}
F_2(X_i) = \sum_{j=1}^{n-1} \sum_{k=1}^{K} T_k(\overrightarrow{P_{ij}P_{i,j+1}})  .
 \end{equation}

It is important to highlight that in certain scenarios where threats cover almost the entire map, rendering the identification of a feasible path becomes challenging, some algorithms may endlessly search for a nonexistent feasible path. To address this, a relatively high-cost value is assigned for passing through a threat center, denoted as $J_{\text{pen}}$. This value is intentionally set to be smaller than infinity, but greater than the cost of any feasible path that could avoid all threats. Therefore, if the output of an algorithm surpasses $J_{\text{pen}}$, it indicates that the path has inevitably crossed a threat or multiple threats, as implied by the multiples of $J_{\text{pen}}$. In our implementation $J_\text{pen} = 1e4$.

%HERE I ONLY COMPARED SEVERAL SOLUTIONS VISUALLY (in terms of crossing threats). Only some algorithms showed a notable difference between $J_{\inf}={e4,e10}$ {Birmin (worse results), EA4eig, EBO, NM (Much better)}
\subsubsection{Altitude Cost}
The second safety criterion pertains to the permissible range of altitude for the UAV's flight. In many applications, it is essential to confine the UAV within designated minimum and maximum altitude values, further influencing path planning. The altitude cost $F_3(X_i)$ is thus calculated on the basis of specified height constraints. Let the minimum and maximum heights be $h_{min}$ and $h_{max}$, respectively. 
The altitude cost for a waypoint $P_{ij}$ unfolds as:
\begin{equation}
H_{ij} = \begin{cases} 
|h_{ij} - \frac{h_{max} + h_{min}}{2}|, & \text{if } h_{min} \leq h_{ij} \leq h_{max} \\
\infty, & \text{otherwise,}
\end{cases}
\end{equation}
where $h_{ij}$ denotes the altitude measured from the waypoint $P_{ij}$ to the terrain's ground, see Fig. \ref{fig:smoothness}.
%the flight height with respect to the terrain's ground.
The total altitude cost for each flight path, considering all waypoints, is expressed as:
\begin{equation}
F_3(X_i) = \sum_{j=1}^{n} H_{ij}.
\end{equation}

\subsubsection{Smoothness Cost}
\begin{figure}
     \centering
    \includegraphics[width=0.6\linewidth]{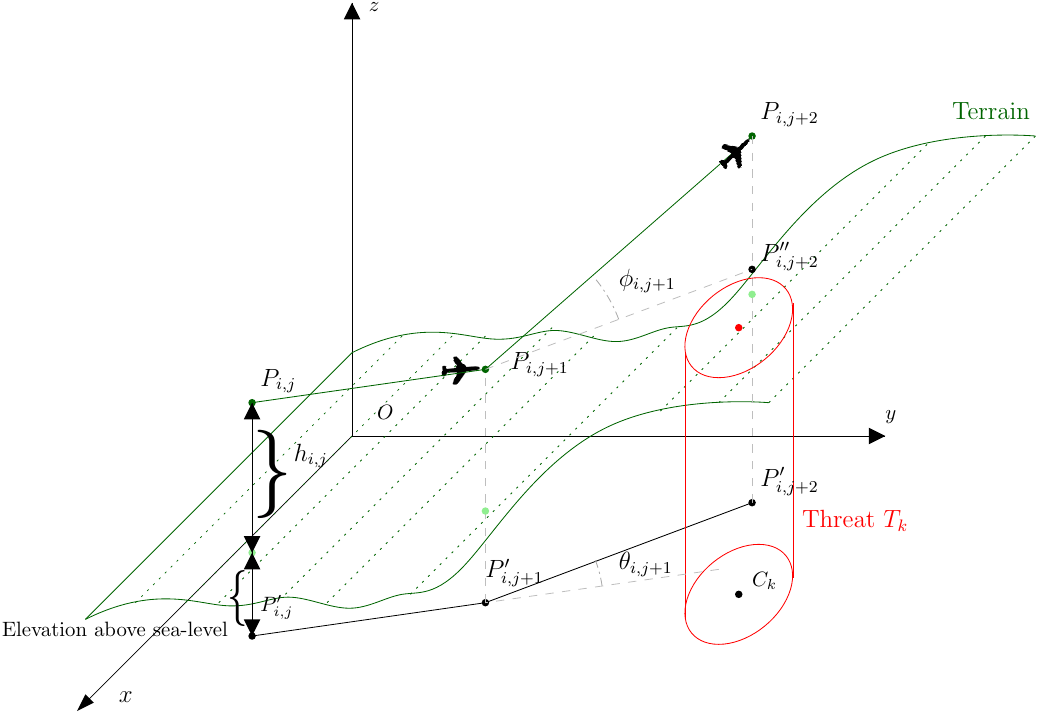}
    \caption{UAV’s turning angle and climbing angle. The dark green points denote the waypoints $P_{ij}$, with their projection on the terrain colored light green, and the projection on the horizontal plane represented by black points $P'_{ij}$. $P''_{i,j+1}$ is the projection of $P_{i,j+1}$ onto a horizontal plane at an altitude of $z_{i,j}$. The red circle represents a potential threat $T_k$.}
    \label{fig:smoothness}
\end{figure}

Besides maintaining an average height, ensuring smooth motion is crucial for both efficiency and safety.
 Smooth paths minimize abrupt turns and altitude changes, which can strain the UAV's dynamics and waste energy. The smoothness cost considers turning (yaw angle) and climbing (pitch angle) angles to regulate motion, promoting smoother trajectories as shown in Fig. \ref{fig:smoothness}.
 
The turning angle ($\theta_{ij}$) between consecutive path segments is calculated as the angle formed by the projection of the segments onto the horizontal plane:
\begin{equation}
\theta_{ij} = \cos^{-1} \left( \frac{\overrightarrow{P'_{i,j-1}P'_{ij}}
\cdot \overrightarrow{P'_{ij}P'_{i,j+1}}}{\| \overrightarrow{P'_{i,j-1}P'_{ij}}
 \| \cdot \| \overrightarrow{P'_{ij}P'_{i,j+1}} \|} \right),
 \end{equation}
similarly, the climbing angle ($\phi_{ij}$) measures the change in altitude between consecutive waypoints:
\begin{equation}
\phi_{ij} = \tan^{-1} \left( \frac{z_{i,j+1} - z_{ij}}{\| \overrightarrow{P'_{ij}P'_{i,j+1}} \|} \right).
\end{equation}

The UAV's trajectory-smoothing cost considers the UAV's turning and climbing angles to generate a trajectory that minimizes large-scale turns and altitude changes, further ensuring smooth UAV movement. The cost model is formulated to penalize excessive turning and climbing angles, as abrupt maneuvers can compromise trajectory efficiency and safety. Specifically, the smoothing cost model calculates the cumulative penalty for deviations from smooth trajectory paths, as dictated by specified penalty coefficients. Thus, the smoothness cost function is defined as:
\begin{equation}
   F_4(X_i) = \beta_1 \sum_{j=2}^{n-1}  \theta_{ij} + \beta_2 \sum_{j=2}^{n-1} |\phi_{ij} - \phi_{i,j-1}|,
\end{equation} 
%\[ J_5(X_n) = \beta_1 \sum_{k=1}^{K-2} \theta_n(k) + \beta_2 \sum_{k=1}^{K-1} |\phi_n(k) - \phi_n(k+1)| \]
here, $\beta_1$ and $\beta_2$ are penalty coefficients for turning and climbing angles, respectively.
%our approach to Smoothness cost considers into turning and climbing rates, indispensable for the formulation of viable paths. The turning angle ($\phi_{ij}$) and climbing angle ($\psi_{ij}$) harmoniously contribute to the smooth cost $F_4(X_i)$, factoring in penalty coefficients $a_1$ and $a_2$.
\subsubsection{Synthesizing the Overall Cost Function}
Taking these four aforementioned costs into account, the path planning algorithm generates trajectories that prioritize smoothness, stability, and safety, ensuring efficient and reliable UAV operations. 

The unification of path-optimality, safety, smoothness, and feasibility constraints coalesces into the overall cost function $F(X_i)$: 
%\[ F(X_i) = \sum_{k=1}^{4} b_kF_k(X_i) \]
\begin{equation}
 F(X_i)=\sum_{k=1}^{4} b_kF_k(X_i),
 \label{F-all}
\end{equation}
where $b_k$ represents the weight coefficient of partial costs. 
%where $b_k$ represents the weight coefficient, and $F_1(X_i)$ to $F_4(X_i)$ correspond to the costs associated with path length, threat avoidance, altitude stability, and smoothness, respectively.
The cost function $F$ is fully determined and serves as the input to the optimization problem in which the objective is to find the path $X^*$ that minimizes $F$, the decision variables are the waypoints $P_{ij} = (x_{ij}, y_{ij}, z_{ij})$ such that $P_{ij} \in O$, where $O$ represents the operating space of the UAVs.

\subsubsection{Limitations of the Problem Definition}
Although the formulation of the objective function is relatively straightforward and natural, we have also identified several limitations and directions that can lead to further improvements. 
\begin{enumerate}
    \item For the height constraint, it is only checked at waypoints, so it might happen that the UAV path will cross a steep hill/ top of a mountain, even when the two waypoints are in the permitted range. This could be addressed by discretizing the path between the waypoints and checking the height constraint, but it would lead to additional computational costs.
    \item Some of the papers use $J_{\text{pen}}= \inf$. We conducted preliminary experiments with this setup and found that in some scenarios (difficult high-density instances) there was no feasible path found for some of the methods (which is more an algorithmic problem, but the ``noninformative'' nature of infinite objective values should be taken into account when designing the optimization formulation).
    %\item for some other scenarios with $J_{\inf}= e^4 or e^5$ the path was crossing threats eventhoug an alterantaive threatless path exist.
\end{enumerate}

\subsection{Exploratory Landscape Analysis of Generated Instances}
We use ELA features to investigate how the selected UAV problems relates to other benchmark suits. For this comparison, we select the BBOB, CEC, and ABS suits that are included in the \directgolib{} library \cite{DIRECTGOLib2023}. In order to calculate the ELA features, we used the flacco library \cite{kerschke2019comprehensive} in R. The ELA feature sets that only need samples of input and function value pairs for the computations were chosen (namely {\tt ela\_distr}, {\tt ela\_meta}, {\tt disp}, {\tt nbc}, {\tt pca}, and {\tt ic}). We used uniform sampling with $250{\times}dim$ samples in $dim=30$ (i.e., $DV = 10$ for the UAV problems). There were 8 problems in the ABS set, 24 problems in the BBOB set, 47 problems in the CEC set, and 56 UAV instances. Before the computation of the ELA features, we used the transformation of function values proposed in \cite{prager2023nullifying}.

For the selection and visualization of the relevant ELA features, we follow the methodology presented in \cite{vskvorc2020understanding}. The features that produced irrelevant, constant, or invalid values were removed. We also removed highly correlated features. After this selection process, the values of the 24 features that remained were normalized. For visualizing the results, we utilized three distinct dimension reduction methods that are commonly used: Principal Component Analysis (PCA), t-Distributed Stochastic Neighbor Embedding (TSNE), and Uniform Manifold Approximation and Projection (UMAP).

\begin{figure}[!t]
    \centering
    \includegraphics[width=\textwidth]{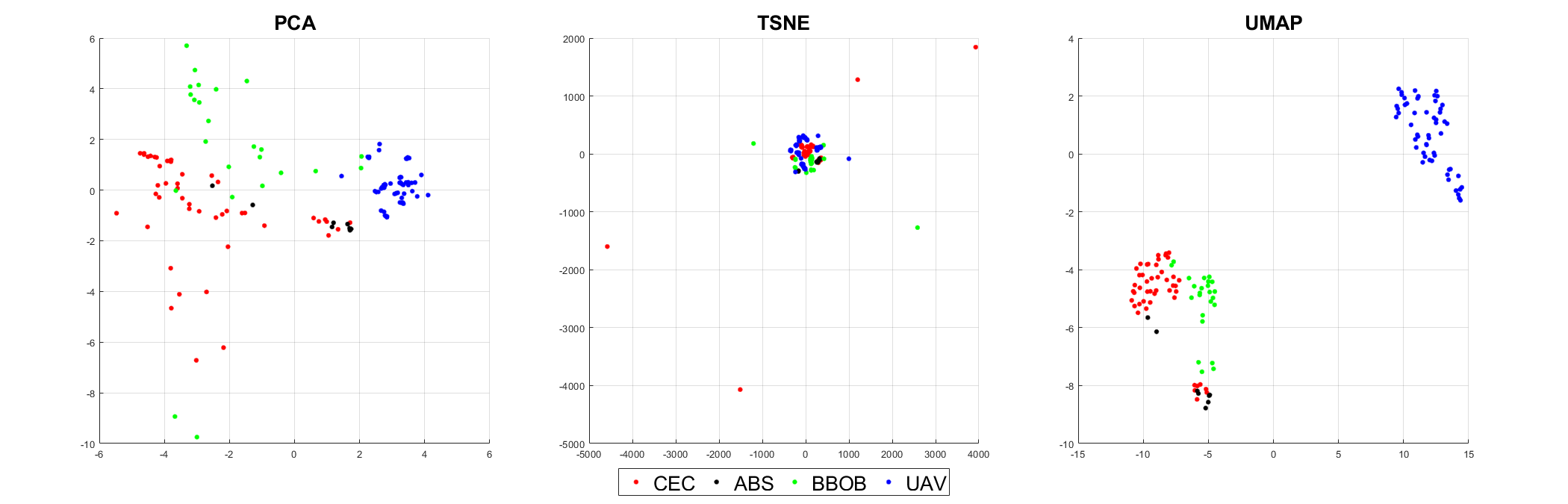}
    \caption{Visualization of the ELA features using different dimension reduction techniques.}
    \label{fig:ela}
\end{figure}

The results of the visualizations are shown in Fig. \ref{fig:ela}. It can readily be seen that although the selected UAV problems are relatively similar to each other (at least in the ELA space), their similarity with the benchmark function in the three sets is very low. This is especially visible from the PCA and UMAP plots. These findings suggest that the selected UAV problems may form an interesting addition to the established benchmark suits, or be a part of newly established real-world problem-based benchmark suits \cite{kudela2022critical}.

\section{Selected Optimization Methods and Experimental Setup}
\label{sec3}
As $F$ is a nonlinear and nonconvex function, solving it using classic methods is not feasible. Instead, various global optimization methods (including both the deterministic and the heuristic techniques) are often used to provide quality solutions in a reasonable amount of time. We have selected 12 well-performing methods from a recent large benchmarking study that are considered state-of-the-art global optimization methods \cite{stripinis2024benchmarking}. For their comparison, we followed the guidelines published in \cite{latorre2021prescription}. For the sake of replicability of our findings \cite{kononova2023importance,back2023evolutionary}, the implementation of the selected methods, the implementation of the optimization model, the instance generation routine, and the implementation of the considered algorithms, as well as the important information about their parametrization, and can be found in a public Zenodo repository \cite{zenodolink}.

%WE generate 28 random terrains from \cite{matlab-terrain} which mimic variyng physical terrains, then for each terrain we randomly distrubute 15/30 threats across the terrain. thus we end up with a 56 sets of terrain that we use for comparison.

%This section describes the classic PSO and its variants including θ-PSO and QPSO which are among the most popular metaheuristic algorithms used for UAV path planning.

\subsection{Brief Description of Selected Methods}
The selected global optimization methods come from four distinct categories. In the first category are the simple local search techniques with restarts. In this category, we selected the Nelder-Mead (\nm) simplex method \cite{nelder1965simplex}.

In the second category are the deterministic optimization methods based on the DIRECT algorithm \cite{jones1993lipschitzian}. Here, we selected two algorithms. The first one was the globally biased version of the DIRECT algorithm called \birmin{} \cite{paulavivcius2020globally}. This algorithm combines the partitioning scheme of the DIRECT algorithm with an interior point local search routine. \birmin{} was also found to have superior performance on the GKLS generator \cite{gaviano2003algorithm, kudela2023computational}. The second algorithm in this category was the two-step Pareto selection-based
DIRECT-type algorithm (\dtcg) \cite{stripinis2022empirical}. Comprehensive numerical investigations have found that \dtcg{} is among the best-performing DIRECT-type methods, particularly for more complex multi-modal problems \cite{stripinis2022directgo, stripinis2024benchmarking}.

In the third category, we selected some of the standard and well-tested evolutionary computation techniques. The first method in this category was Differential Evolution (\de), which is one of the oldest but still widely popular evolutionary computation techniques \cite{storn1997differential, pant2020differential}. At its core, \de{} is a technique that works by maintaining and creating new populations of candidate solutions by using specific combinations of existing solutions and keeping the candidate solution with the best properties for the given optimization problem. Extensive research has been done in extending and hybridizing \de{} \cite{das2010differential,kuudela2023combining}, and in its applications to engineering problems \cite{bujok2023differential}. The second method in this category was Particle Swarm Optimization (\pso), which is also a canonical algorithm \cite{kennedy1995particle}. \pso{} is to this day among the most used and studied evolutionary computation methods \cite{piotrowski2023particle} with an unyielding interest in its possible parametrizations \cite{wang2018particle}. \pso{} was also the method used for the original paper that deals with the UAV planning problem - the variant of \pso{} used in the paper, called \spso, was also added for the comparison.

The last category comprised the top-performing methods from the recent Competitions on single-objective real-parameter numerical optimization at the Congress on Evolutionary Computation (CEC). Many of these methods utilize advanced parameter adaptation schemes, such as Success History Based Adaption (SHBA) and Linear Population Size Reduction (LPSR). The first selected method in this category was the Adaptive Gaining-Sharing Knowledge algorithm (\agsk) \cite{mohamed2020evaluating} - the runner-up in the CEC’20 competition. \agsk{} improved on the original Gaining-Sharing Knowledge algorithm \cite{mohamed2020gaining} by extending the adaptive setting to two control parameters, which control junior and senior gaining and sharing phases during the optimization process. \agsk{} was found as one of the best-performing methods for difficult robotics problems \cite{kuudela2023collection}.

The second method was the hybridization of the \agsk{} algorithm by a \de{} variant called the IMODE
algorithm \cite{sallam2020improved}. This method is denoted as \agskimode. While IMODE won the CEC’20 competition, the hybridized \agskimode{} was fourth in the CEC’21 competition.
The third selected method was \eaeig{} \cite{bujok2022eigen}, which is a hybrid method that uses 3 different algorithms based on \de{} with SHBA, LSPR, and an Eigen transformation approach that is based on the evolution strategy with covariance matrix adaptation (CMA-ES) algorithm \cite{hansen2003reducing}. \eaeig{} was the winner of the CEC’22 competition. The fourth selected method in this category was the Effective Butterfly Optimizer (ebocmar) \cite{kumar2017improving}, which is a swarm-based method that also
used LSPR, SHBA, and a so-called covariance matrix adapted retreat, which uses a strategy similar to CMA-ES algorithm. \ebocmar{} was the winner of the CE’17 competition. The fifth selected method was the Success-history based adaptive differential evolution with linear population size reduction (\lshade) \cite{tanabe2014improving}. This method is based on an adaptive \de, and is perhaps the best-known and widely used example of the successful usage of the SHBA and LPSR schemes. \lshade{} was also found to be very effective in solving complex thermal engineering problems \cite{mauder2024soft, kuudela2024assessment} and for underwater glider path planning \cite{zamuda2019success}.
The last selected method was \elshade, which is a hybridization of the \lshade{} method with the CMA-ES algorithm. \elshade{} was third in the CEC'18 competition.

\subsection{Experimental Setup}
The chosen algorithms are compared over a consistent set of fifty-six generated terrains (see Section \ref{s56}). To study the effect of changing the problem dimensionality, we set up the experiments with a varying number of decision vectors $DV = \{5, 10, 15, 20\}$. Each of the $DV$s, being a triplet $(r,\psi,\phi)$, represents a node on the planned path from start to end, resulting in problems of dimensions $\{15, 30, 45, 60\}$. 
The number of decision vectors directly affects the freedom the path can attain but increases the search space size. The maximum number of function evaluations is set proportionally to the number of decision vectors $max_{evaluation}=3\cdot DV\cdot B$ where $B$ is a base budget. As some of the methods in our comparison utilize parameter control schemes that depend on the remaining available computations (SHBA nad LPSR), we select three base budgets $B=\{ 1e3, 1e4, 1e5\}$. The differences in budgets can also uncover different strengths and weaknesses for the considered methods \cite{piotrowski2025metaheuristics}. For the largest budget, we also imposed a 3 hour limit on the computational time (mainly because the deterministic methods were too costly to run for the entire budget on the largest $DV$ instances). All selected methods started from an identical random seed to ensure a standardized starting point \cite{matousek2022start}. Both the optimization model and implementations of the chosen global optimization algorithms were programmed in MATLAB R2022b and the experiments were run on a workstation with 3.7 GHz AMD Ryzen 9 5900X 12-Core processor and 64 GB RAM.

%% file: Results.tex
\section{Results and Discussion}\label{sec4}

\begin{figure}[!t]
    \centering
    \begin{tabular}{cc}
         \includegraphics[width=0.48\linewidth]{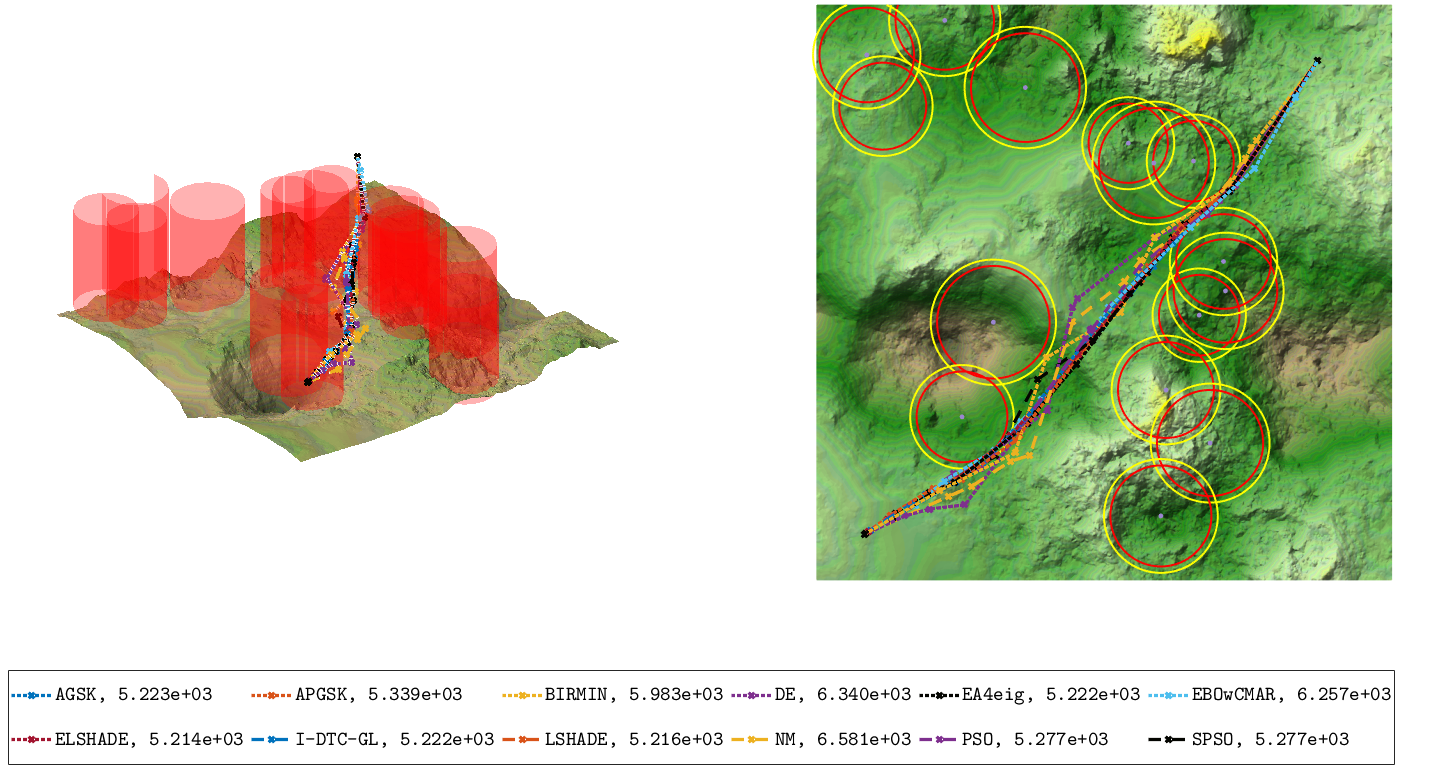}  &
         \includegraphics[width=0.48\linewidth]{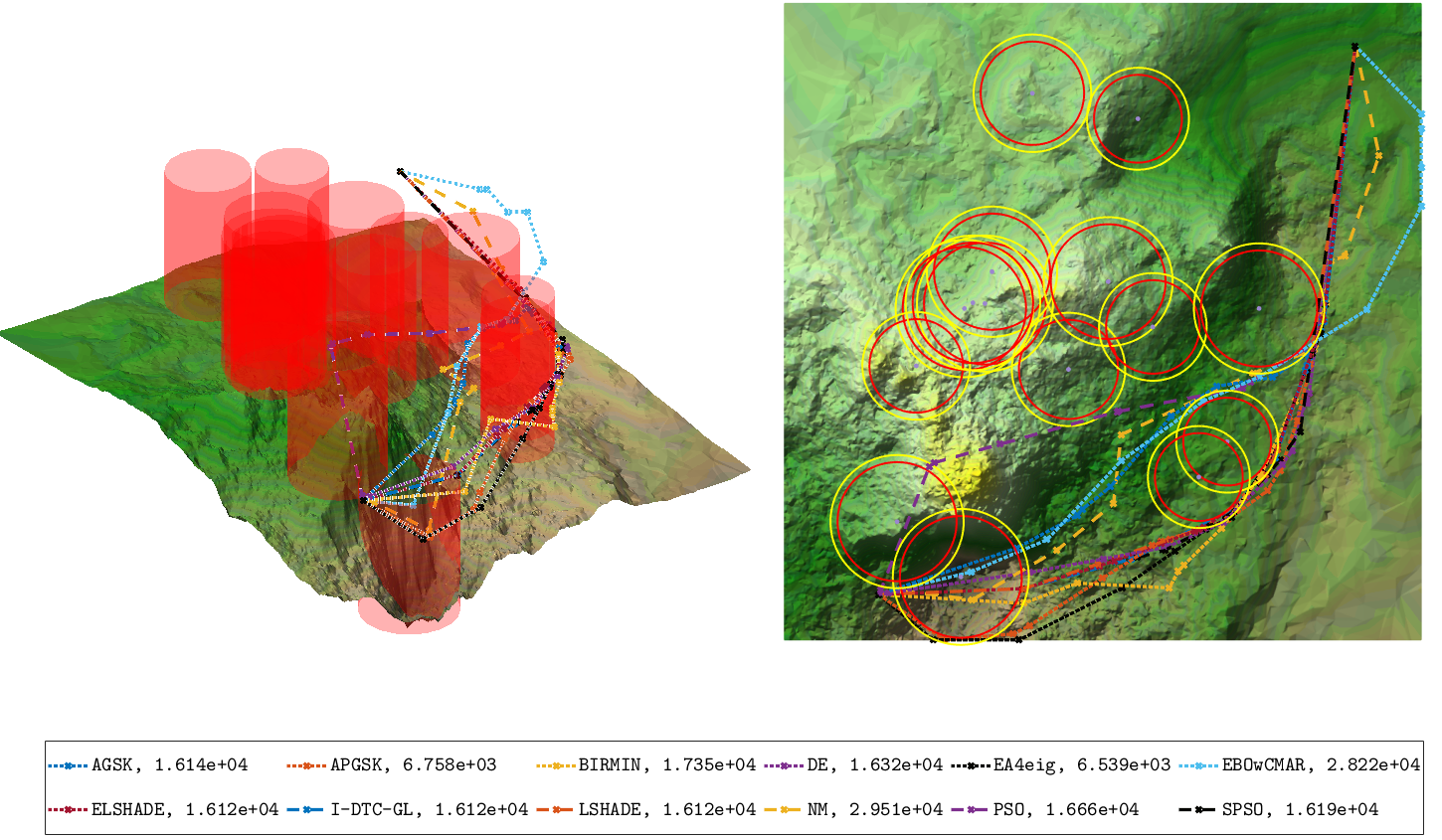}  \\
         a) Terrain 5 & b) Terrain 19 \\
         \includegraphics[width=0.48\linewidth]{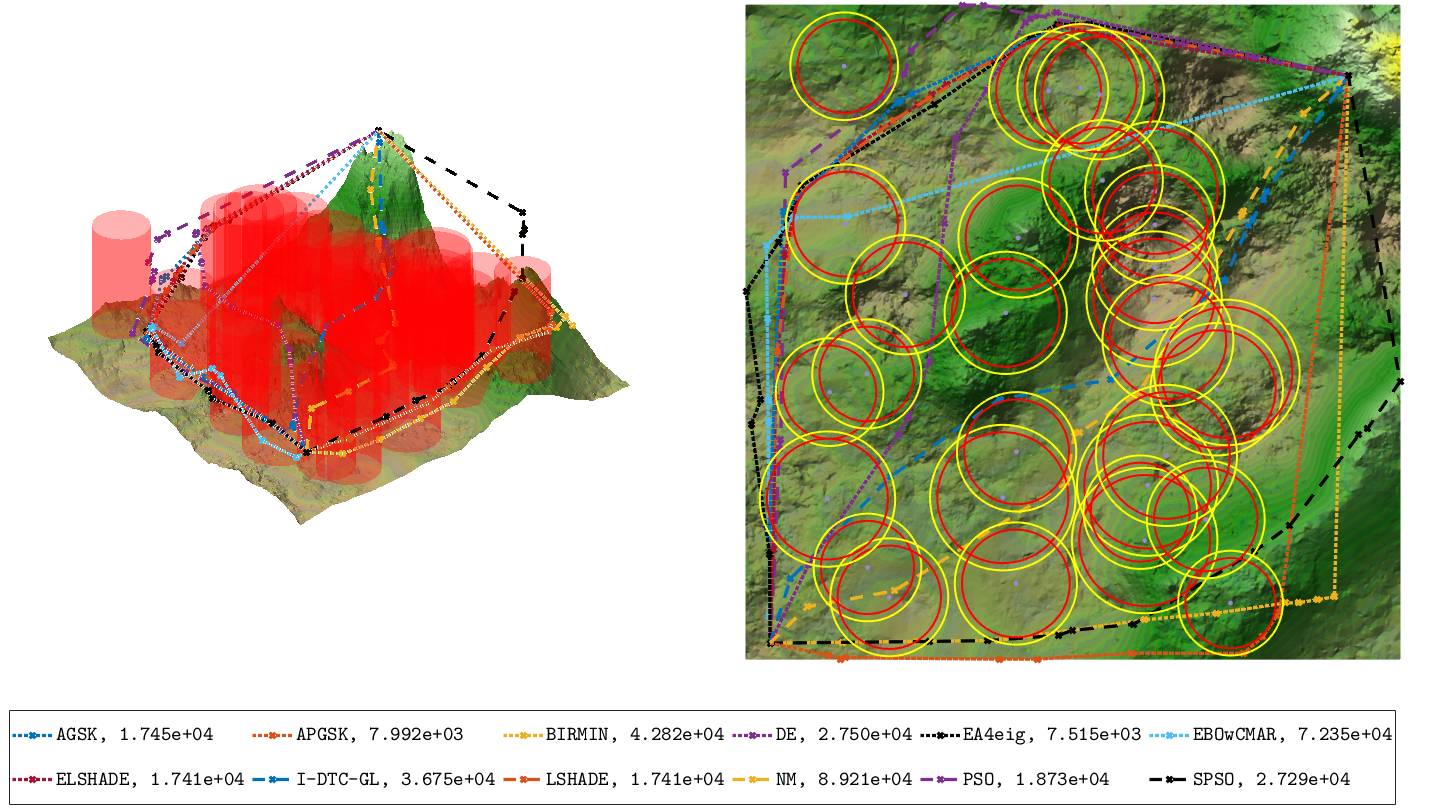}  &
         \includegraphics[width=0.48\linewidth]{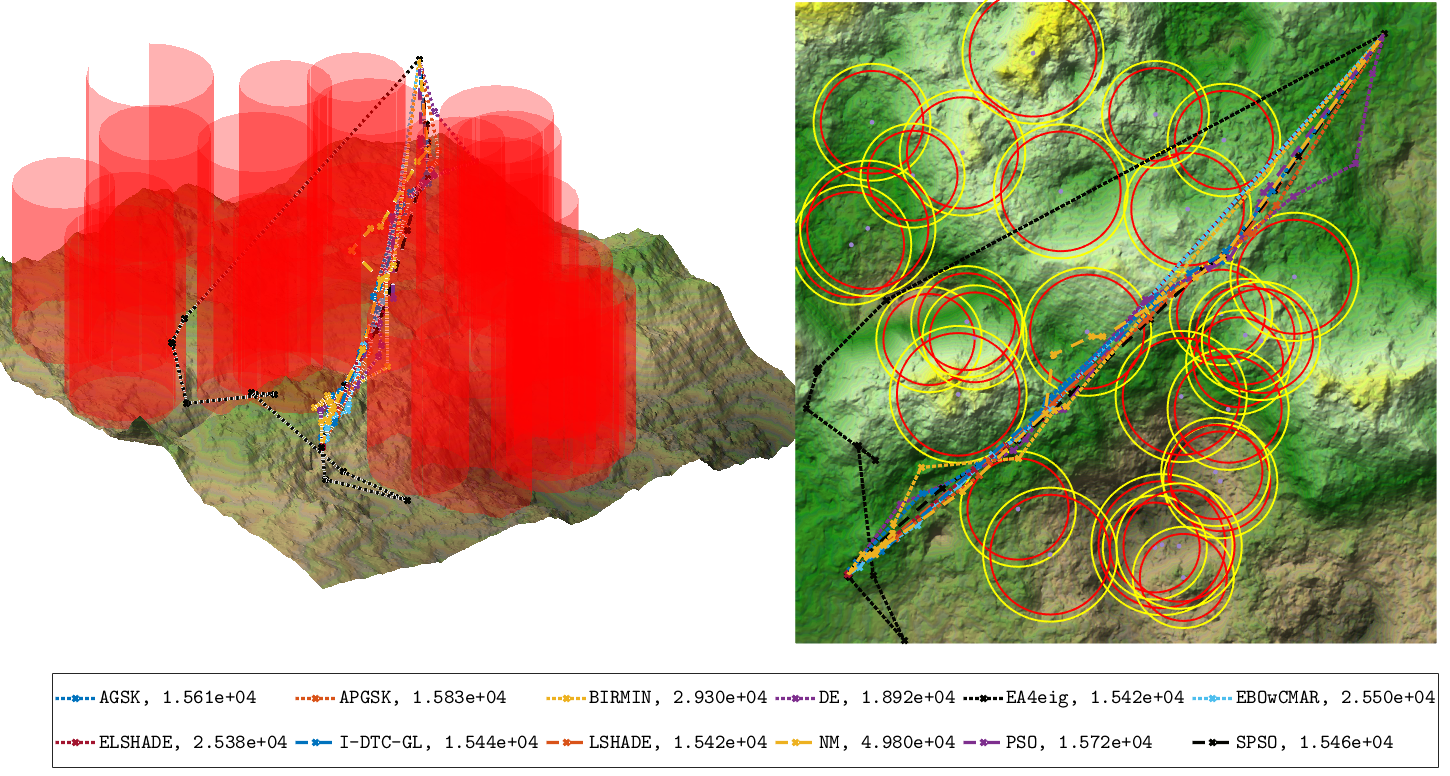}  \\
         c) Terrain 28 & d) Terrain 32 \\
    \end{tabular}
    \caption{Resulting trajectories and objective function values of the different methods in $B = 1e4$, $DV = 10$ setting.}
    \label{fig:trajalg}
\end{figure}

The results of the numerical experiments are best exemplified by Fig.~\ref{fig:trajalg}, which shows the best-found trajectories of the considered methods for some of the problem instances. Here, we can see that there are substantial structural differences between the solutions produced by the considered methods in different settings. In Fig.~\ref{fig:trajalg} a), all the methods were able to find structurally similar solutions,  with the differences in the objective function values signifying various levels of refinement and local search capabilities of these methods. In Fig.~\ref{fig:trajalg} b), only two methods (\agskimode{} and \eaeig) could find a solution that did not violate the constraint of the safety constraint (and hence have an objective value less than $J_{\text{pen}}=1e4$). These two solutions were again structurally similar. In Fig.~\ref{fig:trajalg} c), we also find that only two methods found feasible trajectories (again, \agskimode{} and \eaeig), but this time, these two trajectories were structurally very different - one going through the upper left corner, the other through the lower right corner of the map. Lastly in Fig.~\ref{fig:trajalg} d) we see a situation where none of the methods found any feasible solution, although there definitely exists one (for instance, going through the lower right corner of the map).

\begin{figure}[!t]
    \centering
    \begin{tabular}{cc}
         \includegraphics[width=0.48\linewidth]{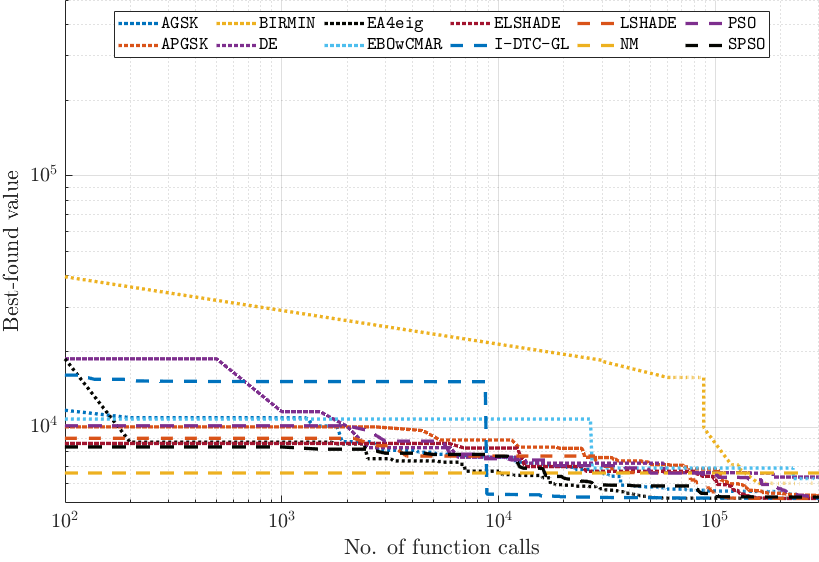}  &
         \includegraphics[width=0.48\linewidth]{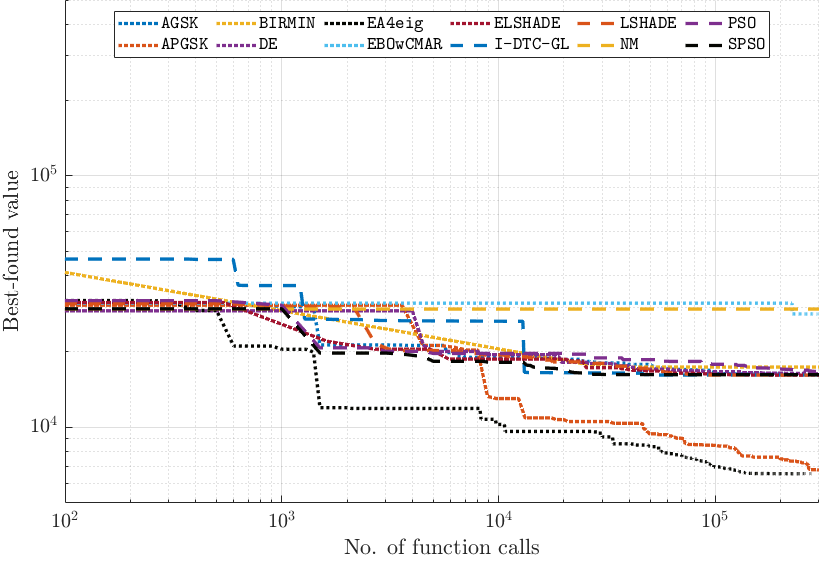}  \\
         a) Terrain 5 & b) Terrain 19 \\
         \includegraphics[width=0.48\linewidth]{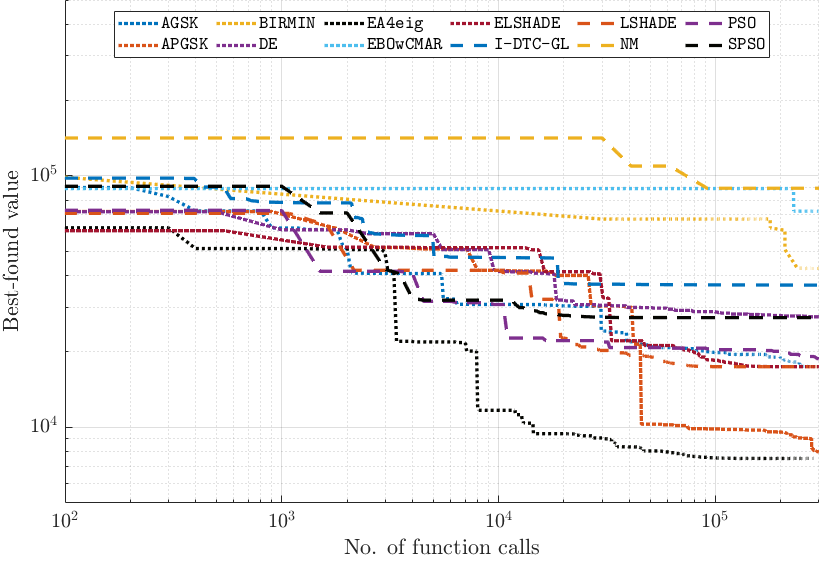}  &
         \includegraphics[width=0.48\linewidth]{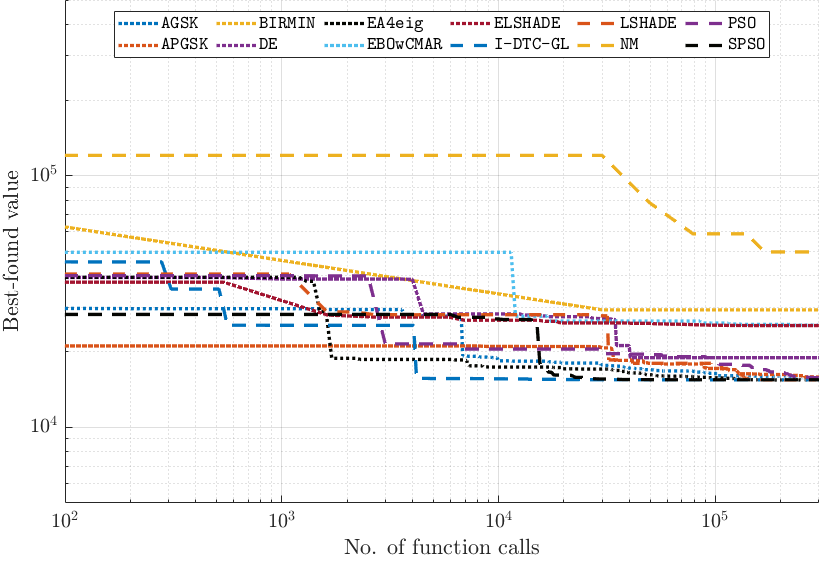}  \\
         c) Terrain 28 & d) Terrain 32 \\
    \end{tabular}
    \caption{Convergence plots of the different methods in 1E4, DV10 setting.}
    \label{fig:conv}
\end{figure}

The convergence plots that correspond to the trajectories in Fig.~\ref{fig:trajalg} are shown in Fig.~\ref{fig:conv}. These display interesting properties of the optimization problem. All the convergence plots show large steps that correspond to the methods finding solutions that went through fewer threads. These steps can appear both after large sections of stagnation, or in relatively short succession after each other. For instances where all methods found feasible trajectories (with objective function values less than $J_\text{pen} = 1e4$), such as Fig.~\ref{fig:conv} a), the feasible solutions were found very soon in the search. Fig.~\ref{fig:conv} b) and c) show that for the problems where only a few methods could find a feasible trajectory the moments of finding them were not predictable - they could happen both towards the very beginning and the very end of the search. Lastly, Fig.~\ref{fig:conv} d) shows that finding ``relatively good'' solutions (that violated just a few of the threads) early on was not a predictor for the success of any of the methods. 

The comparison of the performance of the selected methods will be divided based on the available computation budget and the dimensionality of the problem. As all 56 generated problems have different (and unknown) values of the global optimal solutions, the analysis of the results will be carried out with respect to the best-found solutions across the different methods. As the main metric for the comparison, we use the relative error of the best-found solution of the given method to the best-found solution over all selected methods. For instance, if a method has on a given problem instance a relative error of 0.1, it means that it found a solution with an objective function value 10\% worse than the best method (on the same instance). This metric was chosen because it is independent of the range of the values of the objective function and is comparable over the 56 problem instances. 

\begin{figure}[!t]
    \centering
    \begin{tabular}{cc}
       \includegraphics[width = 0.45\linewidth]{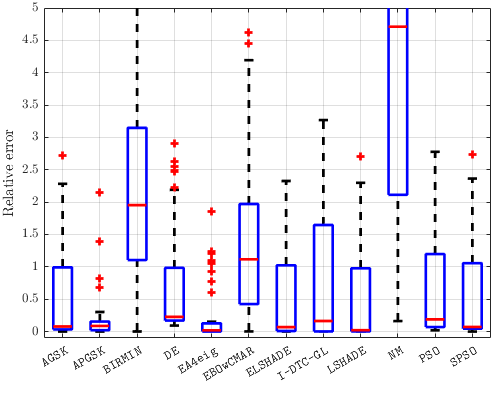}  &  \includegraphics[width = 0.45\linewidth]{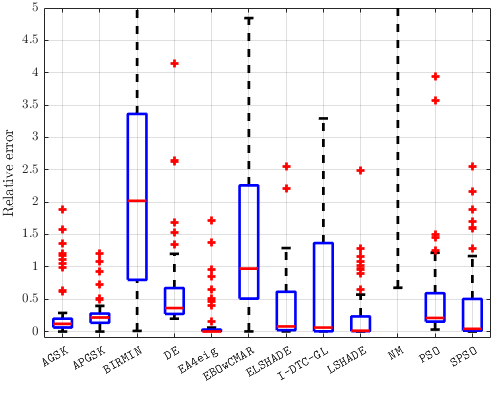}\\
       a) $DV = 5$ & b) $DV = 10$\\[3mm]
       \includegraphics[width = 0.45\linewidth]{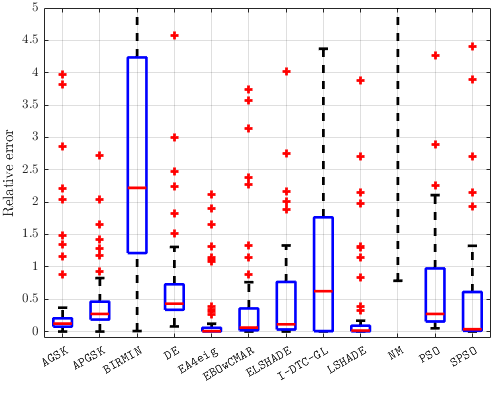} & \includegraphics[width = 0.45\linewidth]{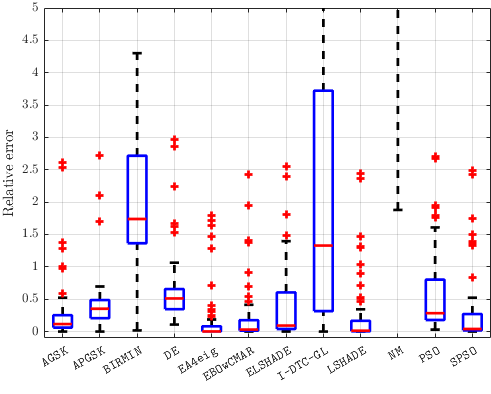}\\
       c) $DV = 15$ & d) $DV = 20$\\
    \end{tabular}
    \caption{Boxplots of the relative errors of the selected methods in different dimensions with $B=1e3$.}
    \label{fig:res1e3}
\end{figure}

The results of the computations for $B=1e3$ and different numbers of $DV$s are summarized in Table~\ref{tab-fe3} and Fig.~\ref{fig:res1e3}. Fig.~\ref{fig:res1e3} shows the boxplots of the relative errors of the selected methods over the 56 instances for different $DV$s.
These boxplots provide valuable information about the distribution of the relative error values for the selected methods. For all values of $DV$ there were several methods for which the median error (the red bar in the boxplot) was very close to 0, meaning that they were able to find a very good solution (close to the best one found by any method) in roughly half of the 56 instances. However, when it comes to the third quantile (the upper part of the blue rectangle), only a few of the methods had it close to 0. In this regard \eaeig{} stands out for all $DV$s, having the boxplot always close to zero. Another noteworthy observation can be made about \ebocmar, for which the distribution of the relative errors gets progressively better with the increase in $DV$. It should be noted that for all considered methods (even the best-performing ones) there were outliers with larger values of relative errors (between 0.5 and 3) and that 
no method could be seen as completely robust in this setting. 

The results summarized in Table~\ref{tab-fe3} show in the respective columns the mean relative error (over the 56 instances, lower is better), the number of found best solutions (out of the 56 instances, higher is better), and the Friedman rank (lower is better) of the selected methods for different $DV$s. The interesting thing to note is that the best methods can be quite different when looking at the three different measures. For $DV=5$, we find that the lowest mean relative error method, \agskimode, was not even in the top 5 methods when looking at the Friedman rank. Conversely, \dtcg, which had a very large portion of the best-found solutions (same as \agskimode), scored a little bit better on the Friedman rank but had a much worse relative mean error. Similar discrepancies between the three measures can also be found for the other $DV$s. On the whole, the one method that stood out as being consistently among the best 3 considered methods for all $DV$s and in all categories was \eaeig. Other notable methods were \agsk{} and \agskimode{} (both having low relative mean errors, signifying robust performance with few outliers), \lshade{} and \spso{} (having low Friedman rank and a large number of wins), and \dtcg{} (having a large number of wins). 

\begin{table}[!t]
\centering
\caption{Summary of the results, $B = 1e3$, the best three methods in the given column are highlighted in bold.}
\label{tab-fe3}
\resizebox{0.99\linewidth}{!}{
\begin{tabular}{l|p{1.2cm}p{1.2cm}p{1.2cm}|p{1.2cm}p{1.2cm}p{1.2cm}|p{1.2cm}p{1.2cm}p{1.2cm}|p{1.2cm}p{1.2cm}p{1.2cm}} \hline 
\multicolumn{1}{l|}{} & \multicolumn{3}{c|}{$DV = 5$} & \multicolumn{3}{c|}{$DV = 10$} & \multicolumn{3}{c|}{$DV = 15$}& \multicolumn{3}{c}{$DV = 20$}\\  
\multicolumn{1}{l|}{} & mean rel. error  & number of wins & Friedman rank        & mean rel. error  & number of wins & Friedman rank        & mean rel. error  & number of wins & Friedman rank        & mean rel. error  & number of wins & Friedman rank        \\ \hline
\agsk           & \textbf{0.407} & 3  & 4.661  & 0.294 & 1  & 5.179  & \textbf{0.452}  & 1  & 5.375  & 0.302  & 3  & 5.196  \\
\agskimode    & \textbf{0.172} & \textbf{11} & 5.321  & \textbf{0.261} & 3  & 6.482  & 0.456  & 3  & 7.018  & 0.432  & 1  & 7.286  \\
\birmin         & 2.590 & 0  & 10.536 & 2.514 & 0  & 10.143 & 3.235  & 0  & 10.786 & 2.931  & 0  & 10.179 \\
\de             & 0.654 & 0  & 8.268  & 0.636 & 0  & 8.625  & 0.785  & 0  & 8.661  & 0.655  & 0  & 8.714  \\
\eaeig         & \textbf{0.235} & \textbf{15} & \textbf{2.643}  & \textbf{0.149} & \textbf{28} & \textbf{2.375}  & \textbf{0.244}  & \textbf{22} & \textbf{2.661}  & \textbf{0.213}  & \textbf{27} & \textbf{2.446}  \\
\ebocmar    & 1.401 & 0  & 10.018 & 1.638 & 0  & 9.125  & 0.782  & 1  & 4.982  & \textbf{0.233}  & 2  & \textbf{3.821}  \\
\elshade & 0.453 & 5  & \textbf{4.232}  & 0.358 & 0  & 4.821  & 0.596  & 1  & 5.804  & 0.397  & 3  & 5.286  \\
\dtcg       & 0.864 & \textbf{11} & 4.821  & 0.700 & \textbf{10} & 5.071  & 1.191  & 8  & 6.161  & 2.195  & 4  & 8.107  \\
\lshade         & 0.408 & 10 & \textbf{3.107}  & \textbf{0.241} & \textbf{10} & \textbf{2.929}  & \textbf{0.423}  & \textbf{11} & \textbf{2.786}  & \textbf{0.258}  & \textbf{11} & \textbf{2.768}  \\
\nm             & 5.537 & 0  & 11.607 & 9.427 & 0  & 11.875 & 13.840 & 0  & 11.929 & 17.554 & 0  & 11.964 \\
\pso            & 0.721 & 0  & 7.232  & 0.513 & 0  & 7.143  & 0.761  & 0  & 7.750  & 0.644  & 0  & 7.839  \\
\spso           & 0.508 & 1  & 5.554  & 0.394 & 4  & \textbf{4.232}  & 0.460  & \textbf{9}  & \textbf{4.089}  & 0.327  & \textbf{5}  & 4.393  \\ \hline
\end{tabular}
}
\end{table}

To perform a more in-depth statistical comparison of the selected algorithms, we followed the guidelines published in \cite{latorre2021prescription}. The results of this comparison are in Table~\ref{tab-pe3}, where we utilized the Wilcoxon signed-rank test to find out whether statistically significant differences exist between the best algorithm (the one with the lowest Friedman rank in the given $DV$) and the other methods. Once the pairwise p-values were computed, we applied the Holm–Bonferroni \cite{holm1979simple} correction method which counteracts the effect of multiple comparisons by controlling the family-wise error rate \cite{aickin1996adjusting}. The only two methods for which no statistically significant difference from the best method (\eaeig) was found were \agskimode{} (for $DV = 5$) and \lshade{} (for $DV = \{5, 15\}$).

\begin{table}[!t]
\centering
\caption{Results of the significance tests, $B = 1e3$. p: p-value computed by the Wilcoxon text p*: p-value corrected with the Holm–Bonferroni procedure. Bold values indicate that statistical
differences do not exist with significance level $\alpha = 0.05$.}
\label{tab-pe3}
\begin{tabular}{l|rr|rr|rr|rr} \hline
      & \multicolumn{2}{c|}{$DV = 5$} & \multicolumn{2}{c|}{$DV = 10$} & \multicolumn{2}{c|}{$DV = 15$} & \multicolumn{2}{c}{$DV = 20$}    \\
       & \multicolumn{1}{l}{p}   & \multicolumn{1}{l|}{p*} & \multicolumn{1}{l}{p}    & \multicolumn{1}{l|}{p*} & \multicolumn{1}{l}{p}    & \multicolumn{1}{l|}{p*} & \multicolumn{1}{l}{p}    & \multicolumn{1}{l}{p*} \\ \hline 
\agsk           & 7.53E-06 & 3.76E-05 & 3.88E-04 & 8.51E-04 & 4.66E-04 & 1.40E-03 & 2.91E-03 & 7.63E-03 \\
\agskimode    & \textbf{6.18E-02} & \textbf{1.24E-01} & 2.84E-04 & 9.37E-04 & 1.26E-04 & 5.05E-04 & 2.76E-05 & 1.65E-04 \\
\birmin         & 7.55E-11 & 8.30E-10 & 2.73E-10 & 2.73E-09 & 7.55E-11 & 8.30E-10 & 7.55E-11 & 8.30E-10 \\
\de             & 7.64E-08 & 5.35E-07 & 4.07E-07 & 2.85E-06 & 6.37E-08 & 5.09E-07 & 4.83E-07 & 3.38E-06 \\
\eaeig         & \textbf{best}     & \textbf{best}     & \textbf{best}     & \textbf{best}     & \textbf{best}     & \textbf{best}     & \textbf{best}     & \textbf{best}     \\
\ebocmar    & 7.55E-11 & 8.30E-10 & 4.60E-10 & 4.14E-09 & 1.54E-05 & 7.69E-05 & 2.54E-03 & 7.63E-03 \\
\elshade & 8.15E-05 & 3.26E-04 & 1.26E-06 & 7.58E-06 & 1.02E-05 & 6.87E-05 & 4.12E-04 & 1.65E-03 \\
\dtcg       & 1.47E-03 & 4.40E-03 & 7.88E-05 & 3.94E-04 & 9.82E-06 & 6.87E-05 & 5.74E-09 & 4.59E-08 \\
\lshade         & \textbf{1.05E-01} & \textbf{1.24E-01} & 1.65E-02 & 1.65E-02 & \textbf{5.03E-02} & \textbf{5.03E-02} & 1.20E-02 & 1.20E-02 \\
\nm             & 7.55E-11 & 7.55E-10 & 7.55E-11 & 8.30E-10 & 7.55E-11 & 8.30E-10 & 7.55E-11 & 8.30E-10 \\
\pso            & 8.88E-11 & 7.10E-10 & 1.24E-08 & 9.95E-08 & 1.08E-08 & 9.69E-08 & 7.97E-11 & 7.55E-10 \\
\spso           & 1.00E-07 & 6.00E-07 & 2.34E-04 & 9.37E-04 & 5.43E-04 & 1.40E-03 & 5.59E-05 & 2.79E-04            \\ \hline 
\end{tabular}
\end{table}

\begin{figure}[!t]
    \centering
    \begin{tabular}{cc}
       \includegraphics[width = 0.45\linewidth]{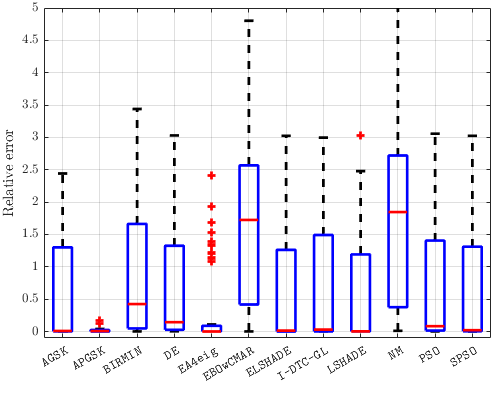}  &  \includegraphics[width = 0.45\linewidth]{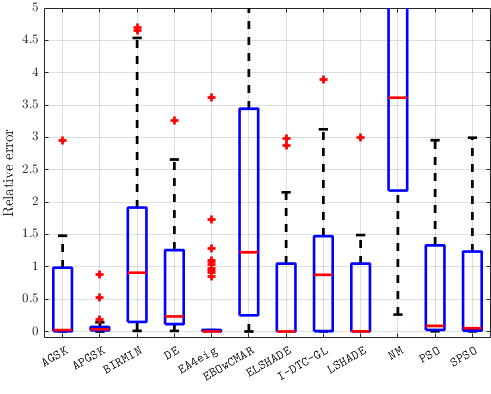}\\
       a) DV = 5 & b) DV = 10\\[3mm]
       \includegraphics[width = 0.45\linewidth]{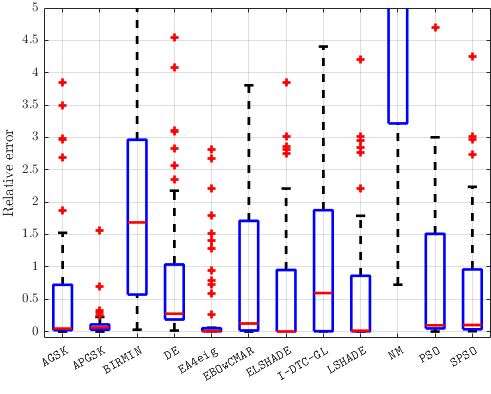} & \includegraphics[width = 0.45\linewidth]{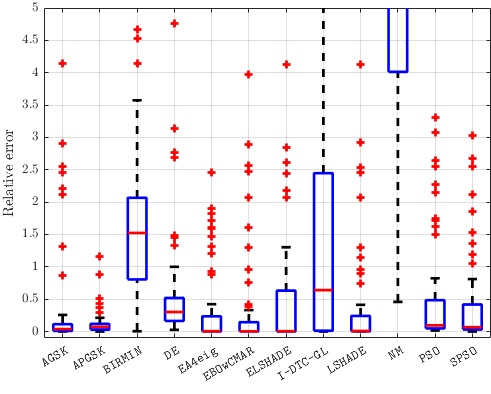}\\
       c) DV = 15 & d) DV = 20\\
    \end{tabular}
    \caption{Boxplots of the relative errors of the selected methods in different dimensions with $B=1e4$.}
    \label{fig:res1e4}
\end{figure}

The results of the computations for $B=1e4$ and different numbers of $DV$s are summarized in Table~\ref{tab-fe4} and Fig.~\ref{fig:res1e4}. Fig.~\ref{fig:res1e4} shows a significant improvement for the deterministic method \birmin{} with the additional allowed number of iterations. The list of top three gainers from the extended computations also includes \agskimode{} and \nm{} (a more detailed analysis of the gains the methods got from the increase in $B$ will be investigated in later text and is summarized in Table \ref{perdev}).
%%%%%%
% (((Not sure if it is worth stressing))): Fig.~\ref{fig:res1e4} exclusively presents boxplots of the relative errors of the selected methods (remember it is a ratio of the solution found by a certain method in comparison to the best one found with same search space size and computational budget). That is why methods with marginal improvement over $n_{max}=4$  exhibit bigger boxplots in contrast to the agile methods which have smaller relative error and thus are represented by smaller boxplots.
%These boxplots provide valuable information about the distribution of the relative error values for the selected methods. For all values of $DV$ there were several methods for which the median error (the red bar in the boxplot) was very close to 0, meaning that they were able to find 
In contrast to the results for $B=1e3$, here the median error gets closer to 0 (between 0 and 0.2) for 60\% of the considered methods for all values of $DV$. This indicates a substantial improvement that is parallelly reflected in many methods, meaning that they were able to find a very good solution (close to the best one found by any method) in roughly half of the 56 instances. However, only two methods have the third quantile (the upper part of the blue rectangle) consistently close to 0. In all $DV$s, \eaeig{} and \agskimode{} have the entire boxplot always below 0.45 (in most cases even below 0.2). That is, they were able to find a good solution (close to the best one found by other methods) in all of the 56 instances, with the exception of a few outliers. For \agskimode, even these outliers had relatively low values, especially for $DV=\{5,10\}$.
%Having an average improvement from $1e3$ to $1e4$  of 28\% \ref{perdev}. \agskimode{} improved almost all its solutions, resulting in even fewer outliers compared to \eaeig{} \ref{fig:res1e4}.%with boxplots laying entirely under a relative error of 0.25 and 

Despite demonstrating such improved robustness, with small numbers of outliers and worst relative errors not exceeding 1.6, and attaining a good number of best solutions \agskimode{} still cannot achieve a low Friedman rank, not even among the top 3 (low-scoring) methods, in any dimension (shown in Table~\ref{tab-fe4}). With respect to the number of wins and the Friedman rank, \eaeig{} performed best at the small search space of $DV = 5$. 
Although \eaeig{} substantially outperformed \elshade{} in $B=1e3$, %and has a notable improvement given a larger budget 8.7\% , slightly falling behind \elshade{} 11.3\%, \ref{perdev}.
\elshade{} can find a higher number of optimal paths in $DV =\{10,15,20\}$ and ranks as best method in $DV =\{10,15\}$ according to Friedman ranks. However, even for these larger search spaces, the difference between these two methods is not statistically significant (as shown in Table~\ref{tab-pe4}).
%, and \eaeig{} maintains a low relative error, allowing it to accept the null hypothesis of being similar to the best method in most cases \ref{tab-pe4}.
Upon observing \eaeig{} we notice a condensed area of outliers roughly around $ \left[1,1.5 \right]$ (for $DV = \{5, 10\}$), imposing questions about the nature of these outliers. Looking deeply at each of the outliers' solutions we have found that \eaeig{} and \agskimode{} have structurally very similar solutions with small shifts of \eaeig{} barely passing a threat (or multiple threats), which adds an additional penalty $J_{\text{pen}} = 1e4$.
% (we can add figures here of multiple solutions to further support the similarity between \eaeig{} and \agskimode{}).
%not sure if this should be mentioned or left out (again it is a statement that is verified for two cases, outliers of $DV = \{5, 10\}$, but this does not hold true for $DV = 20$).
\ebocmar{} kept the good constant progress of the relative errors with the increase in $DV$ and impressively had the lowest Friedman rank for $DV =20$. Other notable methods were \agsk{} and \agskimode{} (both having low relative mean errors, signifying robust performance with few outliers) and \lshade{} (having low Friedman rank and a noteworthy number of wins). For the $B=1e4$ base budget, the best-performing methods predominantly include \eaeig, \elshade, \lshade, \agskimode, and \agsk.

\begin{table}[]
\centering
\caption{Summary of the results, $B = 1e4$, the best three methods in the given column are highlighted in bold.}
\label{tab-fe4}
\resizebox{0.99\linewidth}{!}{
\begin{tabular}{l|p{1.2cm}p{1.2cm}p{1.2cm}|p{1.2cm}p{1.2cm}p{1.2cm}|p{1.2cm}p{1.2cm}p{1.2cm}|p{1.2cm}p{1.2cm}p{1.2cm}} \hline 
\multicolumn{1}{l|}{} & \multicolumn{3}{c|}{$DV = 5$} & \multicolumn{3}{c|}{$DV = 10$} & \multicolumn{3}{c|}{$DV = 15$}& \multicolumn{3}{c}{$DV = 20$}\\  
\multicolumn{1}{l|}{} & mean rel. error  & number of wins & Friedman rank        & mean rel. error  & number of wins & Friedman rank        & mean rel. error  & number of wins & Friedman rank        & mean rel. error  & number of wins & Friedman rank        \\ \hline
\agsk           & \textbf{0.481} & 6  & 4.625  & \textbf{0.399} & 4  & 4.643  & \textbf{0.523} & 3  & 5.304  & 0.467 & 5  & 5.375  \\
\agskimode    & \textbf{0.018} & \textbf{13} & 4.429  & \textbf{0.069} & \textbf{10} & 4.982  & \textbf{0.114} & \textbf{9}  & 5.446  & \textbf{0.129} & 5  & 6.232  \\
\birmin         & 0.967 & 0  & 9.429  & 1.268 & 0  & 9.786  & 2.004 & 0  & 10.589 & 1.833 & 0  & 10.250 \\
\de             & 0.589 & 0  & 8.161  & 0.647 & 0  & 8.607  & 0.801 & 0  & 8.804  & 0.669 & 0  & 8.786  \\
\eaeig         & \textbf{0.337} & \textbf{17} & \textbf{2.857}  & \textbf{0.278} & \textbf{15} & \textbf{2.911}  & \textbf{0.320} & \textbf{10} & \textbf{3.089}  & \textbf{0.331} & \textbf{11} & \textbf{3.732}  \\
\ebocmar    & 1.865 & 0  & 11.036 & 2.148 & 0  & 9.679  & 0.898 & 0  & 6.000  & \textbf{0.369} & \textbf{10} & \textbf{2.964}  \\
\elshade & 0.620 & 7  & \textbf{3.768}  & 0.513 & \textbf{19} & \textbf{2.821}  & 0.611 & \textbf{23} & \textbf{2.911}  & 0.521 & \textbf{18} & \textbf{3.214}  \\
\dtcg       & 0.855 & 3  & 5.446  & 0.856 & 0  & 5.911  & 1.177 & 5  & 5.732  & 1.495 & 1  & 7.554  \\
\lshade         & 0.485 & \textbf{8}  & \textbf{3.375}  & 0.489 & 8  & \textbf{3.179}  & 0.662 & 5  & \textbf{3.839}  & 0.491 & 5  & 3.839  \\
\nm             & 1.826 & 0  & 11.250 & 4.274 & 0  & 11.857 & 7.107 & 0  & 11.982 & 9.670 & 0  & 11.946 \\
\pso            & 0.802 & 0  & 7.857  & 0.680 & 0  & 7.268  & 0.888 & 1  & 7.625  & 0.703 & 0  & 7.375  \\
\spso           & 0.673 & 2  & 5.768  & 0.623 & 0  & 6.357  & 0.718 & 0  & 6.679  & 0.596 & 1  & 6.732          \\ \hline 
\end{tabular}
}
\end{table}

\begin{table}[]
\centering
\caption{Results of the significance tests, $B = 1e4$. p: p-value computed by the Wilcoxon text p*: p-value corrected with the Holm–Bonferroni procedure. Bold values indicate that statistical
differences do not exist with significance level $\alpha = 0.05$.}
\label{tab-pe4}
\begin{tabular}{l|rr|rr|rr|rr} \hline
      & \multicolumn{2}{c|}{$DV = 5$} & \multicolumn{2}{c|}{$DV = 10$} & \multicolumn{2}{c|}{$DV = 15$} & \multicolumn{2}{c}{$DV = 20$}    \\
       & \multicolumn{1}{l}{p}   & \multicolumn{1}{l|}{p*} & \multicolumn{1}{l}{p}    & \multicolumn{1}{l|}{p*} & \multicolumn{1}{l}{p}    & \multicolumn{1}{l|}{p*} & \multicolumn{1}{l}{p}    & \multicolumn{1}{l}{p*} \\ \hline 
\agsk           & 3.22E-04 & 1.29E-03 & 1.97E-02 & \textbf{7.86E-02} & 2.54E-02 & \textbf{7.62E-02} & 9.34E-05 & 4.67E-04 \\
\agskimode    & \textbf{7.82E-01} & \textbf{7.82E-01} & \textbf{2.47E-01} & \textbf{4.93E-01} & \textbf{4.43E-01} & \textbf{8.86E-01} & 3.75E-02 & \textbf{1.50E-01} \\
\birmin         & 1.16E-09 & 9.31E-09 & 7.55E-11 & 8.30E-10 & 7.55E-11 & 8.30E-10 & 1.52E-10 & 1.52E-09 \\
\de             & 2.14E-06 & 1.24E-05 & 1.02E-05 & 5.10E-05 & 6.97E-06 & 4.88E-05 & 2.88E-09 & 1.73E-08 \\
\eaeig         & \textbf{best}     & \textbf{best}     & \textbf{5.90E-01} & \textbf{5.90E-01} & \textbf{7.75E-01} & \textbf{8.86E-01} & \textbf{6.83E-01} & \textbf{9.76E-01} \\
\ebocmar    & 5.38E-10 & 5.38E-09 & 3.74E-10 & 3.36E-09 & 5.40E-05 & 3.24E-04 & \textbf{best}     & \textbf{best}     \\
\elshade & 7.64E-03 & 2.29E-02 & \textbf{best}     & \textbf{best}     & \textbf{best}     & \textbf{best}     & \textbf{4.88E-01} & \textbf{9.76E-01} \\
\dtcg       & 2.06E-06 & 1.24E-05 & 1.55E-06 & 9.31E-06 & 4.00E-04 & 2.00E-03 & 8.13E-10 & 7.32E-09 \\
\lshade         & 3.75E-02 & \textbf{7.50E-02} & \textbf{7.02E-02} & \textbf{2.10E-01} & 1.39E-03 & 5.54E-03 & \textbf{9.29E-02} & \textbf{2.79E-01} \\
\nm             & 3.19E-10 & 3.51E-09 & 1.16E-10 & 1.16E-09 & 7.55E-11 & 8.30E-10 & 7.55E-11 & 8.30E-10 \\
\pso            & 6.97E-10 & 6.27E-09 & 3.01E-07 & 2.11E-06 & 9.47E-07 & 7.57E-06 & 1.22E-09 & 8.57E-09 \\
\spso           & 5.81E-08 & 4.07E-07 & 1.31E-07 & 1.05E-06 & 1.31E-07 & 1.18E-06 & 1.05E-09 & 8.41E-09      \\ \hline         
\end{tabular}
\end{table}

\begin{figure}[!t]
    \centering
    \begin{tabular}{cc}
       \includegraphics[width = 0.45\linewidth]{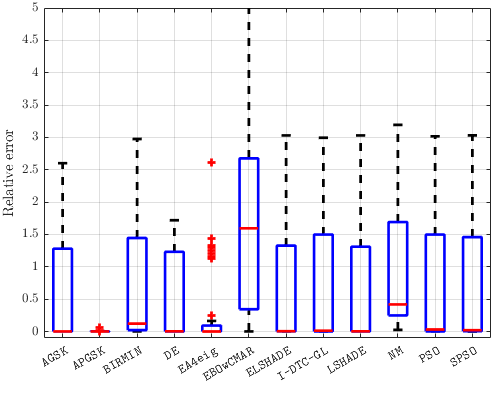}  &  \includegraphics[width = 0.45\linewidth]{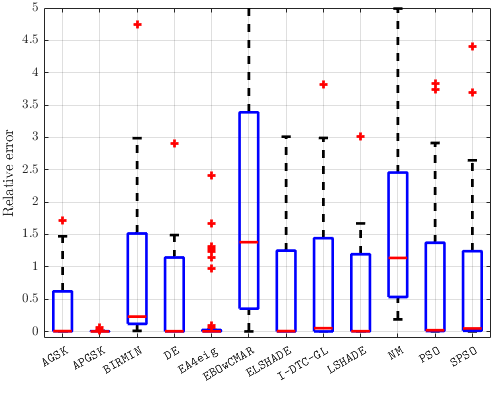}\\
       a) DV = 5 & b) DV = 10\\[3mm]
       \includegraphics[width = 0.45\linewidth]{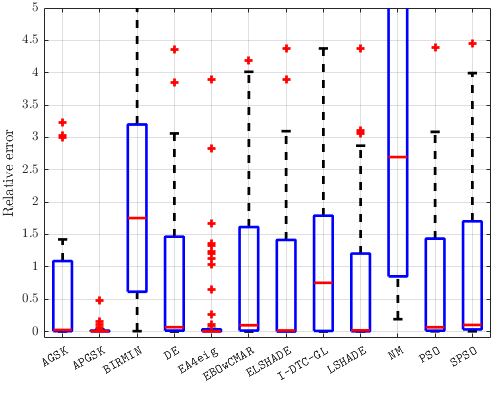} & \includegraphics[width = 0.45\linewidth]{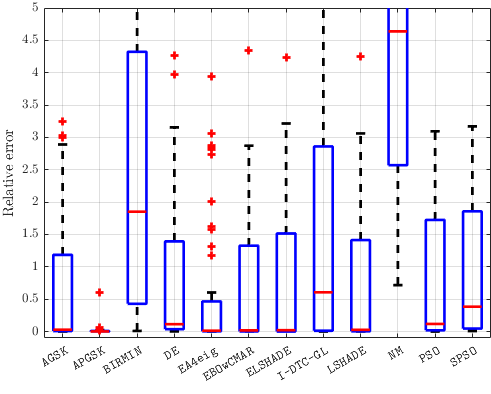}\\
       c) DV = 15 & d) DV = 20\\
    \end{tabular}
    \caption{Boxplots of the relative errors of the selected methods in different dimensions with $B=1e5$.}
    \label{fig:res1e5}
\end{figure}

The results of the computations for the largest base budget $B=1e5$ (a maximum of one hundred thousand function calls capped by three hours of computation) and different numbers of $DV$s are summarized in Table~\ref{tab-fe5} and Fig.~\ref{fig:res1e5}. Fig.~\ref{fig:res1e5} shows an interesting development. For $DV=5$, we find that the majority of the considered methods (\agsk, \birmin, \elshade, \dtcg, \lshade, \pso, \spso) have a very similar distribution of relative errors. The main exceptions are \agskimode{} and \eaeig, which perform much better than the rest, and \ebocmar, which has the highest relative errors. This is also somewhat true for the other $DV$s, with the main difference being the progressively worsening performance of \birmin, \dtcg, and \nm. 

Table~\ref{tab-fe5} also uncovers some interesting insights. For $DV=5$, the three best methods in all categories are \agskimode, \eaeig, and surprisingly, the ``standard'' \de. In larger $DV$s, \de{} gets replaced with \agsk, \elshade, and \ebocmar, but still keeps a relatively good performance. These results suggest that with a large enough computational budget, the differences in performance of many of the considered methods become very low. However, two methods stood out - \agskimode{} and \eaeig{} have consistently the lowest Friedman ranks, and the difference between them and the rest is statistically significant, as shown in Table~\ref{tab-pe5}.

\begin{table}[!ht]
\centering
\caption{Summary of the results, $B = 1e5$, the best three methods in the given column are highlighted in bold.}
\label{tab-fe5}
\resizebox{0.99\linewidth}{!}{
\begin{tabular}{l|p{1.2cm}p{1.2cm}p{1.2cm}|p{1.2cm}p{1.2cm}p{1.2cm}|p{1.2cm}p{1.2cm}p{1.2cm}|p{1.2cm}p{1.2cm}p{1.2cm}} \hline 
\multicolumn{1}{l|}{} & \multicolumn{3}{c|}{$DV = 5$} & \multicolumn{3}{c|}{$DV = 10$} & \multicolumn{3}{c|}{$DV = 15$}& \multicolumn{3}{c}{$DV = 20$}\\   
\multicolumn{1}{l|}{} & mean rel. error  & number of wins & Friedman rank        & mean rel. error  & number of wins & Friedman rank        & mean rel. error  & number of wins & Friedman rank        & mean rel. error  & number of wins & Friedman rank        \\ \hline
\agsk           & 0.496 & 9  & 4.018  & \textbf{0.339} & \textbf{8}  & 4.554  & \textbf{0.508} & 3  & 4.411  & \textbf{0.720} & 3  & 5.321  \\
\agskimode    & \textbf{0.004} & \textbf{13} & \textbf{3.571}  & \textbf{0.006} & \textbf{18} & \textbf{3.464}  & \textbf{0.022} & \textbf{15} & \textbf{3.375}  & \textbf{0.019} & \textbf{21} & \textbf{3.089}  \\
\birmin         & 0.687 & 0  & 9.214  & 0.794 & 0  & 9.821  & 2.267 & 0  & 10.821 & 2.564 & 0  & 10.589 \\
\de             & \textbf{0.482} & \textbf{12} & \textbf{3.670}  & 0.407 & 5  & 4.393  & 0.815 & 1  & 6.464  & 0.846 & 0  & 7.411  \\
\eaeig         & \textbf{0.263} & \textbf{17} & \textbf{3.313}  & \textbf{0.231} & 6  & \textbf{3.857}  & \textbf{0.304} & \textbf{9}  & \textbf{3.446}  & \textbf{0.521} & 3  & 3.821  \\
\ebocmar    & 1.736 & 0  & 11.438 & 2.083 & 0  & 10.839 & 0.996 & 2  & 7.571  & 0.761 & \textbf{5}  & \textbf{3.554}  \\
\elshade & 0.639 & 3  & 5.509  & 0.536 & \textbf{13} & \textbf{3.929}  & 0.920 & \textbf{18} & \textbf{3.777}  & 0.950 & \textbf{18} & \textbf{3.714}  \\
\dtcg       & 0.870 & 0  & 6.357  & 0.761 & 2  & 6.089  & 1.169 & 2  & 5.946  & 1.639 & 4  & 6.786  \\
\lshade         & 0.556 & 2  & 4.982  & 0.441 & 4  & 4.143  & 0.768 & 6  & 4.759  & 0.862 & 1  & 5.161  \\
\nm         & 1.023 & 0  & 10.500 & 1.903 & 0  & 11.321 & 3.934 & 0  & 11.643 & 6.075 & 0  & 11.964 \\
\pso            & 0.877 & 0  & 7.464  & 0.689 & 0  & 7.482  & 0.913 & 0  & 7.214  & 1.082 & 1  & 7.339  \\
\spso           & 0.810 & 0  & 7.964  & 0.578 & 0  & 8.107  & 1.015 & 0  & 8.571  & 1.184 & 0  & 9.250           \\ \hline 
\end{tabular}
}
\end{table}

\begin{table}[!ht]
\centering
\caption{Results of the significance tests, $B = 1e5$. p: p-value computed by the Wilcoxon text p*: p-value corrected with the Holm–Bonferroni procedure. Bold values indicate that statistical
differences do not exist with significance level $\alpha = 0.05$.}
\label{tab-pe5}
\begin{tabular}{l|rr|rr|rr|rr} \hline
      & \multicolumn{2}{c|}{$DV = 5$} & \multicolumn{2}{c|}{$DV = 10$} & \multicolumn{2}{c|}{$DV = 15$} & \multicolumn{2}{c}{$DV = 20$}    \\
       & \multicolumn{1}{l}{p}   & \multicolumn{1}{l|}{p*} & \multicolumn{1}{l}{p}    & \multicolumn{1}{l|}{p*} & \multicolumn{1}{l}{p}    & \multicolumn{1}{l|}{p*} & \multicolumn{1}{l}{p}    & \multicolumn{1}{l}{p*} \\ \hline 
\agsk           & 3.06E-02 & \textbf{9.19E-02} & 4.30E-03 & 1.29E-02 & 3.07E-05 & 1.23E-04 & 3.43E-07 & 1.71E-06 \\
\agskimode    & \textbf{3.00E-01} & \textbf{3.00E-01} & \textbf{best}     & \textbf{best}     & \textbf{best}     & \textbf{best}     & \textbf{best}     & \textbf{best}     \\
\birmin         & 1.73E-08 & 1.56E-07 & 7.55E-11 & 8.30E-10 & 7.55E-11 & 8.30E-10 & 7.55E-11 & 8.30E-10 \\
\de             & \textbf{7.03E-02} & \textbf{1.41E-01} & 2.91E-03 & 1.16E-02 & 2.91E-08 & 1.75E-07 & 2.73E-10 & 2.18E-09 \\
\eaeig         & \textbf{best}     & \textbf{best}     & \textbf{8.98E-02} & \textbf{8.98E-02} & \textbf{2.40E-01} & \textbf{2.40E-01} & 2.16E-03 & 6.49E-03 \\
\ebocmar    & 1.11E-10 & 1.22E-09 & 7.55E-11 & 8.30E-10 & 2.03E-09 & 1.62E-08 & 4.65E-03 & 6.13E-03 \\
\elshade & 1.14E-05 & 5.71E-05 & 1.20E-03 & 6.01E-03 & 8.62E-03 & 1.72E-02 & 3.07E-03 & 6.49E-03 \\
\dtcg       & 7.37E-07 & 4.42E-06 & 8.36E-07 & 5.01E-06 & 4.83E-07 & 2.42E-06 & 1.95E-07 & 1.17E-06 \\
\lshade         & 8.54E-04 & 3.42E-03 & 4.77E-03 & 1.29E-02 & 6.70E-04 & 2.01E-03 & 1.02E-05 & 4.08E-05 \\
\nm             & 2.20E-08 & 1.76E-07 & 7.55E-11 & 7.55E-10 & 7.55E-11 & 8.30E-10 & 7.55E-11 & 8.30E-10 \\
\pso            & 2.65E-07 & 1.85E-06 & 4.62E-08 & 3.24E-07 & 1.03E-08 & 7.19E-08 & 3.03E-09 & 2.12E-08 \\
\spso           & 2.36E-09 & 2.36E-08 & 1.11E-09 & 8.85E-09 & 1.60E-10 & 1.44E-09 & 7.55E-11 & 7.55E-10     \\ \hline         
\end{tabular}
\end{table}

\begin{figure}
    \centering
    \begin{tabular}{cc}
         \includegraphics[width=0.48\linewidth]{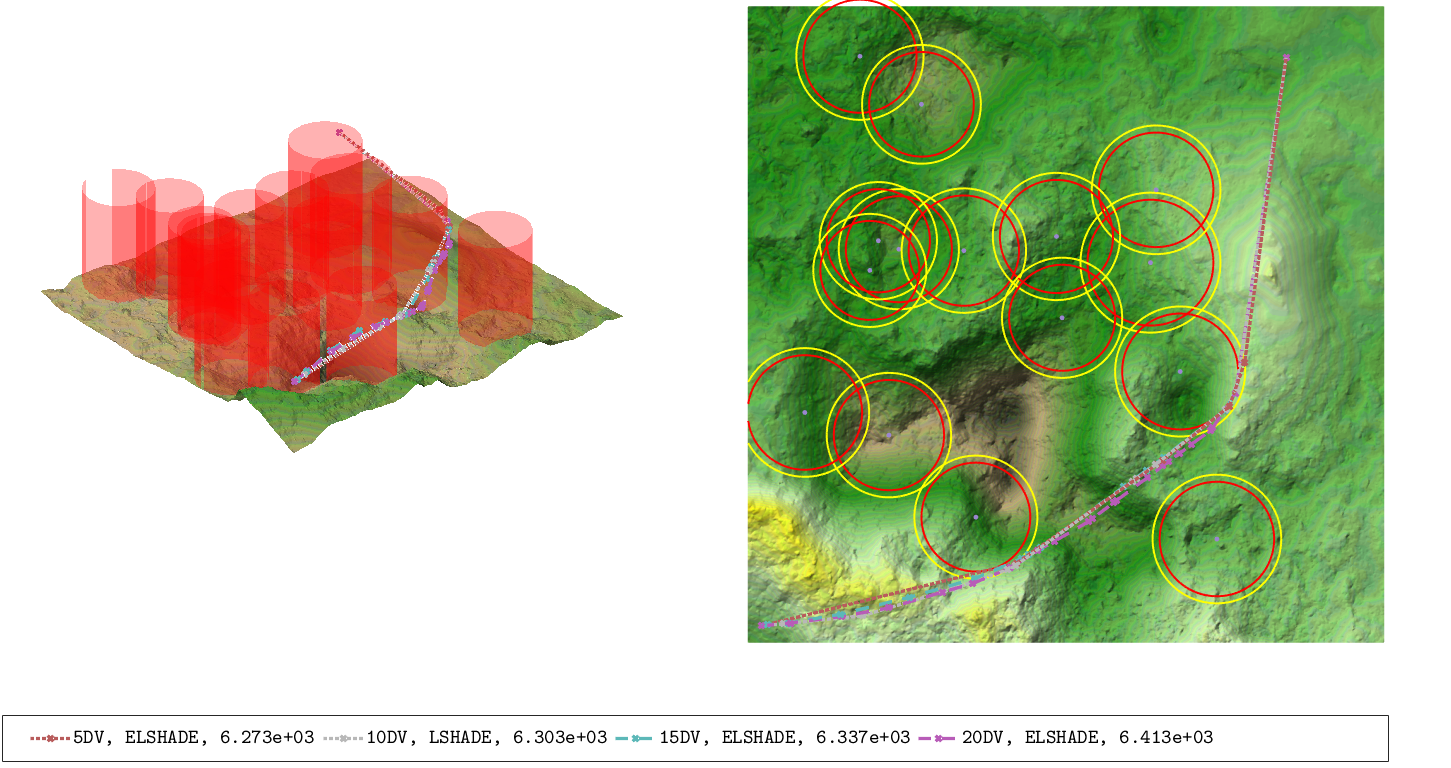}  &
         \includegraphics[width=0.48\linewidth]{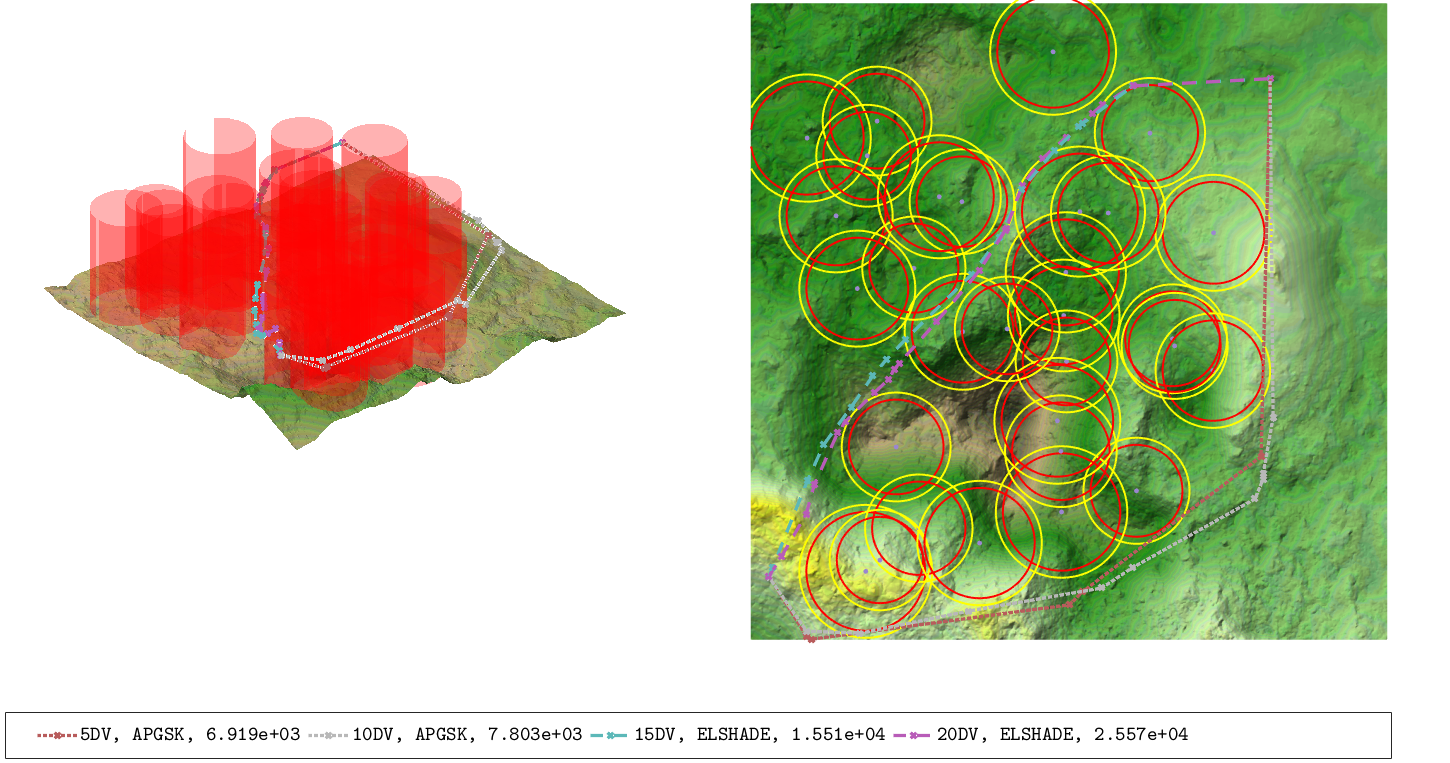}  \\
         a) Terrain 1 & b) Terrain 2 \\
         \includegraphics[width=0.48\linewidth]{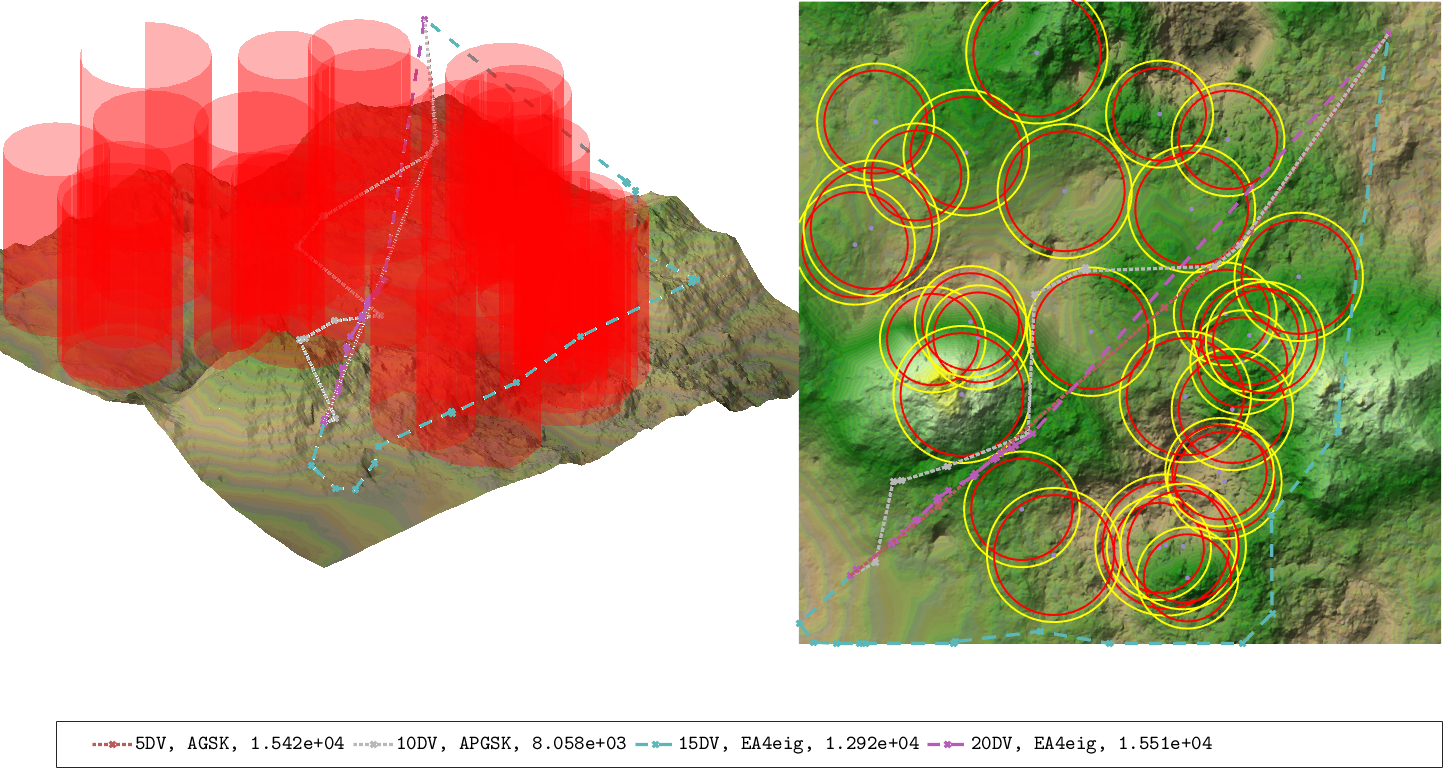}  &
         \includegraphics[width=0.48\linewidth]{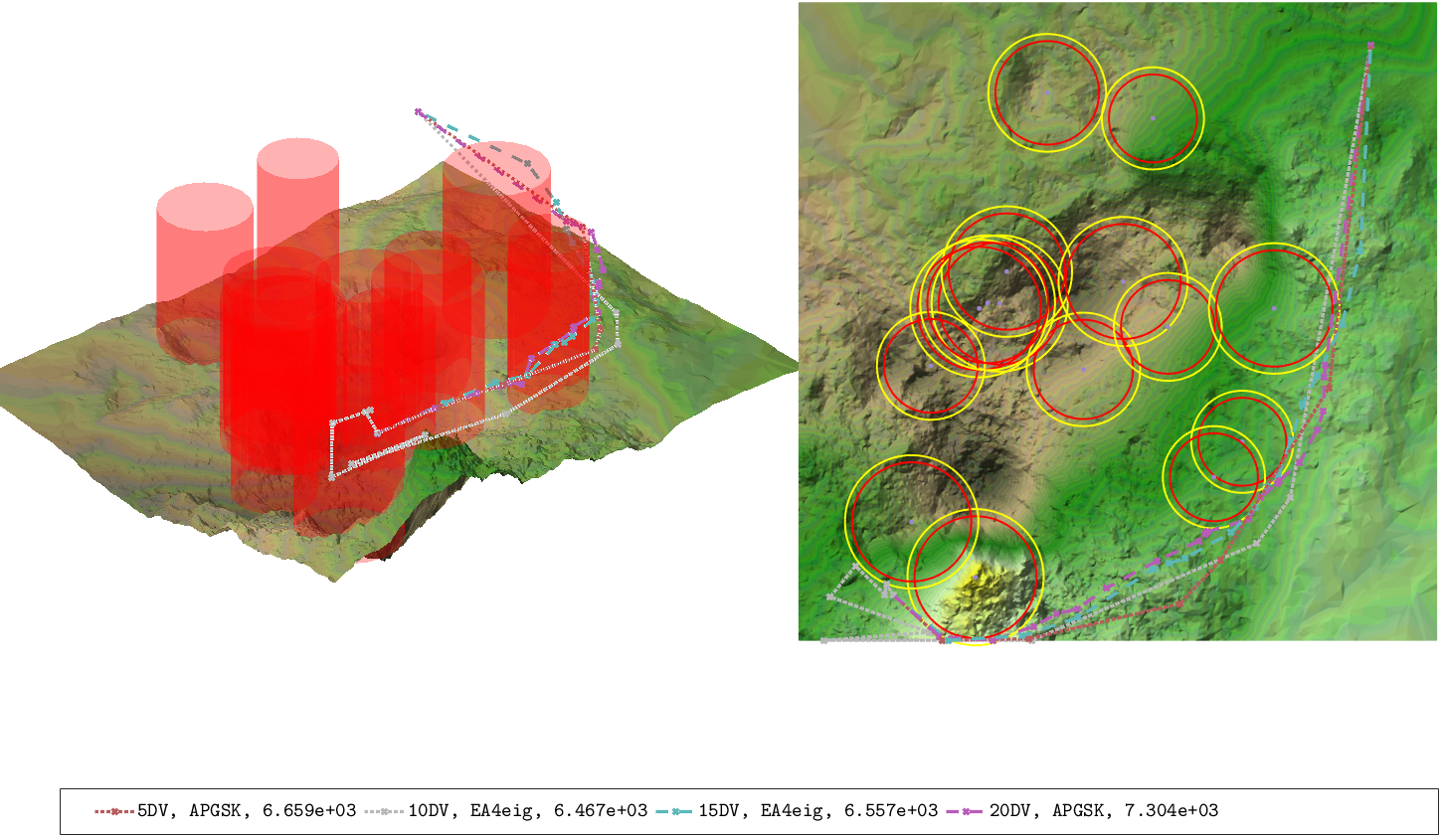}  \\
         c) Terrain 8 & d) Terrain 9 \\
          \includegraphics[width=0.48\linewidth]{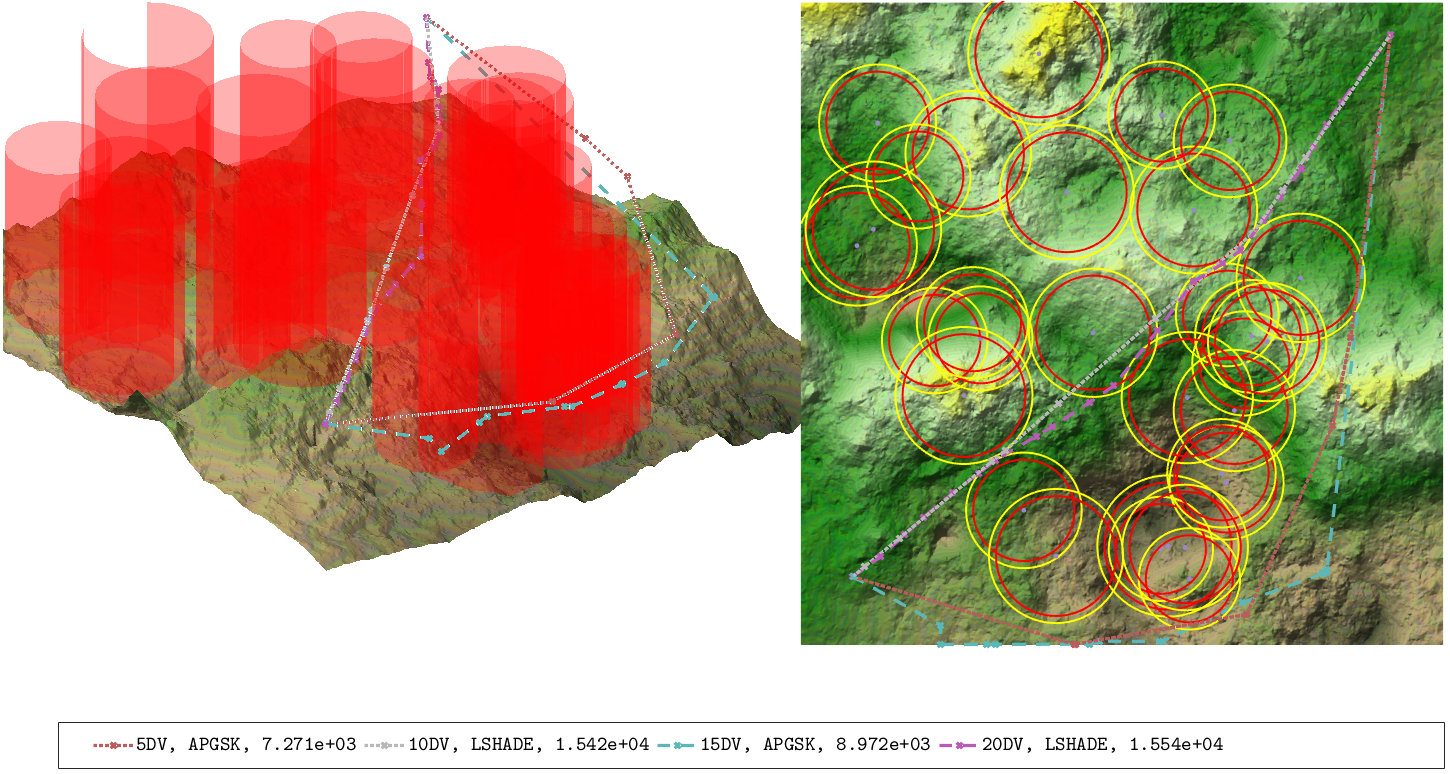}  &
         \includegraphics[width=0.48\linewidth]{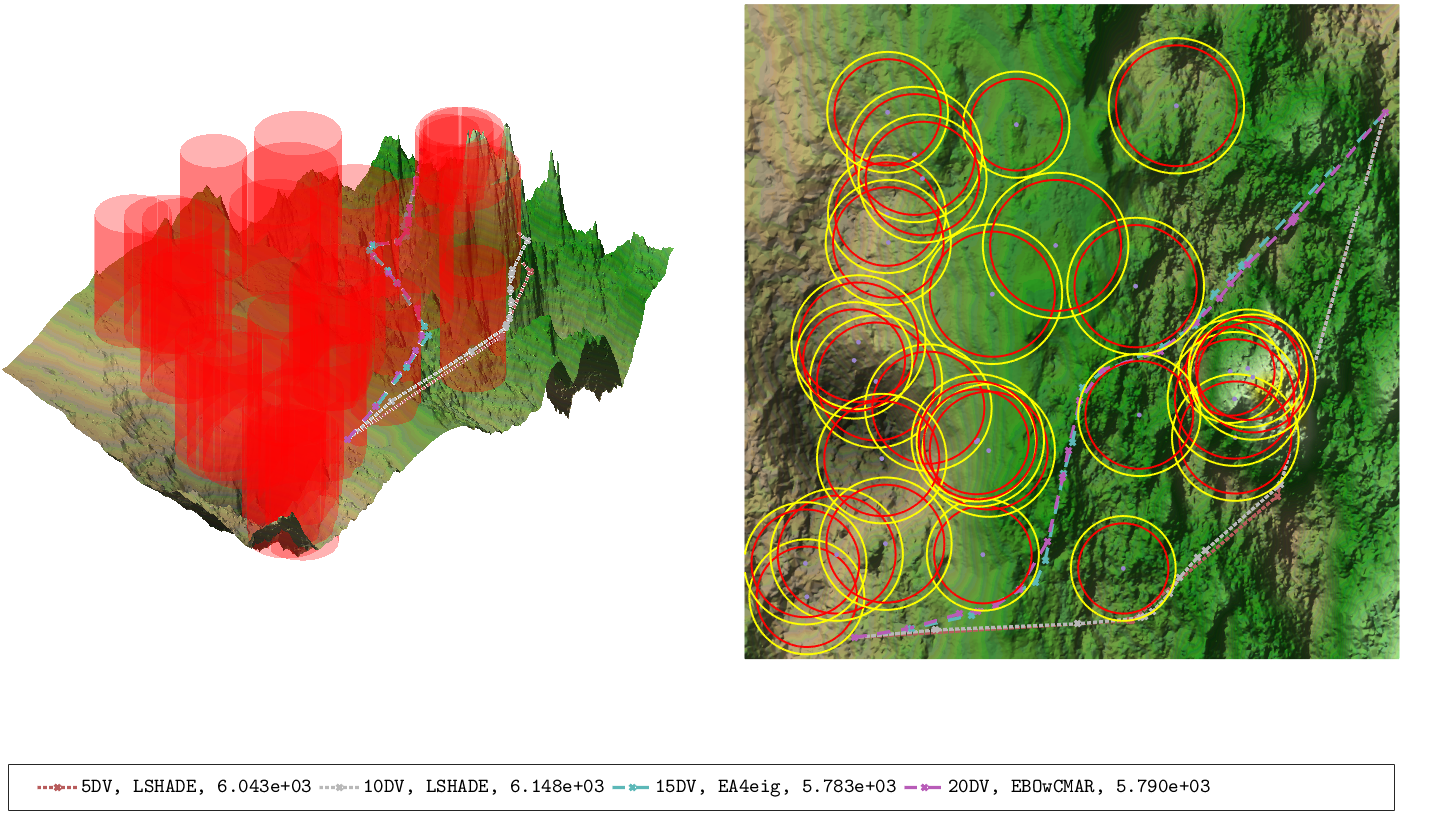}  \\
         e) Terrain 32 & f) Terrain 50 \\
    \end{tabular}
    \caption{The trajectories and objective function values of the best methods in different dimensions, $B=1e4$.}
    \label{fig:trajdim}
\end{figure}

\begin{table}[]
\caption{Data for the investigation of the variable dimension characteristic of the UAV problem.}
\begin{tabular}{lllllllll} \hline 
      & \multicolumn{2}{c}{$DV = 5$} & \multicolumn{2}{c}{$DV = 10$} & \multicolumn{2}{c}{$DV = 15$} & \multicolumn{2}{c}{$DV = 20$}  \\
      & mean   & number  & mean   & number & mean   & number & mean   & number 
      \\ $B$ & rel. error & of wins & rel. error &  of wins & rel. error & of wins & rel. error & of wins     \\ \hline 
$1e3$ & 0.008          & 45             & 0.136          & 7              & 0.292          & 3              & 0.524          & 1              \\
$1e4$ & 0.020          & 50             & 0.106          & 5              & 0.237          & 1              & 0.579          & 0              \\
$1e5$/3h & 0.005          & 51             & 0.155          & 1              & 0.127          & 2              & 0.222          & 2             \\ \hline 
\end{tabular}
\label{tab-vd}
\end{table}

Next, we investigate the ``variable dimension'' characteristic of the studied problem. The 56 instances were the same across the different $DV$s, giving us the opportunity to find the ``best dimension'' ($DV$) for this problem. To study this, we identified the best solutions for each of the 56 scenarios for each of the 4 $DV$ values. Fig.~\ref{fig:trajdim} shows the resulting trajectories for a handful of selected scenarios that cover (in their structure) most of the remaining results. In Fig.~\ref{fig:trajdim} a) and d) the trajectories in the different dimensions look very similar and the only difference lies in their ``refinement'' (the best one for a) is in $DV=5$, while the best one for d) is in $DV = 10$). In Fig.~\ref{fig:trajdim} b) and e) there are two good trajectories (in $DV = 5$ and $DV = 10$ for b), and in $DV = 5$ and $DV = 15$ for e)), while the other two go through some of the threats. In Fig.~\ref{fig:trajdim} c) there was only a single good trajectory (in $DV = 10$). Finally, in Fig.~\ref{fig:trajdim} f) all four trajectories were good (not passing through any threats), but they were not similar as in Fig.~\ref{fig:trajdim} a). 

Further insights into the variable dimension characteristic of the studied problem can be found in Table~\ref{tab-vd}. Here, we find that overall, it is more beneficial to solve the studied UAV problems in lower dimensions (especially in $DV=5$). Even though the increase in $DV$ can in theory bring additional refinements to the trajectories (and, hence, lower objective values), the added computational costs and difficulty of the problem that comes from the dimension increase are too big. It would be interesting to further study the possibility of developing variable-dimension methods that start in lower $DV$ and, once they find a feasible solution, refine it in higher $DV$s.

%\begin{table}[]
%\caption{Average relative improvement in the best objective function value (over all methods) from the increased computational budget.}
% \begin{tabular}{lllll} \hline 
%  $B$ increase             & $DV = 5$   & $DV = 10$  & $DV = 15$  & $DV = 20$  \\ \hline 
% from $1e3$ to $1e4$ & 0.096 & 0.110 & 0.112 & 0.071 \\
% from $1e3$ to $1e5$ & 0.114 & 0.111 & 0.167 & 0.196 \\ \hline 
% \end{tabular}
% \end{table}
\begin{table}[]
\caption{Average relative improvement in the best objective function value (over all methods) from the increased computational budget.}
\begin{tabular}{llllll}
\hline
$B$ increase                         & \begin{tabular}[c]{@{}c@{}}Density of \\ Threats\end{tabular} & $DV = 5$    & $DV = 10$ & $DV = 15$ & $DV = 20$ \\ \hline
\multirow{2}{*}{from $1e3$ to $1e4$} & low                                                           & 0.02802 & 0.0317    & 0.06238   & 0.03321   \\ 
                                     & high                                                          & 0.16386    & 0.18926   & 0.16214   & 0.10825   \\ \hline
\multirow{2}{*}{from $1e3$ to $1e5$} & low                                                           & 0.03479    & 0.03576   & 0.07391   & 0.05737   \\ 
                                     & high                                                          & 0.19306    & 0.18539   & 0.25998   & 0.33442   \\ \hline
\end{tabular}
\label{tab-avr-rel}
\end{table}

%comparing different budgets
Having all three budgets discussed separately, Table~\ref{tab-avr-rel} depicts the average improvement for all methods with an increased number of function calls. It distinguishes between two sets of problems: those with sparse threats and those with denser threats, where finding a feasible solution is quite a challenging task by itself. Table~\ref{tab-avr-rel} reveals that the solutions in $DV=\{5,10\}$ improve significantly with the increased allocated computational budget from $1e3$ to $1e4$, while the additional budget of $1e5$ does not yield significant improvements. Conversely, in $DV=\{15,20\}$, solutions improve mainly within the extended budget of $B = 1e5$. This is particularly surprising since, for $DV=20$, some methods needed more than three hours of computation and therefore were truncated far before reaching the $B = 1e5$ call limit (meaning that the function calls are actually less than $B = 1e5$). Nonetheless, they still performed better than they did for $B = 1e4$.
It can also be seen that the high-density problems show better improvement over $B = 1e5$, in contrast to low-density ones. This is mainly because methods can quickly reach a near-optimal solution on low-density problems (thus the improvement is minimal), while they require more time to skip local minima and find better solutions on high-density problems.
% shall I comment previous sentence since we do not have to give reasons to all results, do we?
% (not sure if you want to include the following) Statically speaking we see that that the number of outliers decreases as we have more function calls and the boxplots can better represent the best found solutions (just a notice that may be redundant, feel free to further comment on it or delete it).
% Considering \ref{perdev}, which represent how methods evolve with more allocated budget, we find some insightful results. The top three "progressors" are  \nm, \birmin, and \agskimode. (has been mentioned before at line 219, The list of top three .....).

\begin{table}[!ht]
\caption{Average relative improvement in the best objective function value from the increased computational budget for all methods.}
\label{perdev}
\begin{tabular}{llcrrrr}\hline
Method                    & $B$ increase                         & \multicolumn{1}{l}{\begin{tabular}[c]{@{}l@{}}Density of \\ Threats\end{tabular}} & \multicolumn{1}{l}{$DV = 5$} & \multicolumn{1}{l}{$DV = 10$} & \multicolumn{1}{l}{$DV = 15$} & \multicolumn{1}{l}{$DV = 20$} \\ \hline
\multirow{4}{*}{\agsk}     & \multirow{2}{*}{from $1e3$ to $1e4$} & Low                                                                               & 11.08\%                      & 10.29\%                       & 12.85\%                       & 10.33\%                       \\
                          &                                      & High                                                                              & 5.89\%                       & 11.56\%                       & 10.13\%                       & 3.94\%                        \\
                          & \multirow{2}{*}{from $1e3$ to $1e5$} & Low                                                                               & 11.86\%                      & 11.87\%                       & 15.48\%                       & 15.72\%                       \\
                          &                                      & High                                                                              & 9.19\%                       & 14.55\%                       & 24.74\%                       & 13.78\%                       \\ \hline
\multirow{4}{*}{\agskimode}    & \multirow{2}{*}{from $1e3$ to $1e4$} & Low                                                                               & 15.33\%                      & 17.65\%                       & 26.41\%                       & 31.90\%                       \\
                          &                                      & High                                                                              & 33.88\%                      & 35.97\%                       & 39.71\%                       & 24.88\%                       \\
                          & \multirow{2}{*}{from $1e3$ to $1e5$} & Low                                                                               & 16.93\%                      & 21.38\%                       & 30.77\%                       & 37.25\%                       \\
                          &                                      & High                                                                              & 38.36\%                      & 38.17\%                       & 55.36\%                       & 56.09\%                       \\ \hline
\multirow{4}{*}{\birmin}   & \multirow{2}{*}{from $1e3$ to $1e4$} & Low                                                                               & 41.12\%                      & 38.07\%                       & 37.48\%                       & 19.73\%                       \\
                          &                                      & High                                                                              & 66.30\%                      & 52.83\%                       & 36.52\%                       & 46.89\%                       \\
                          & \multirow{2}{*}{from $1e3$ to $1e5$} & Low                                                                               & 52.40\%                      & 39.22\%                       & 38.43\%                       & 26.19\%                       \\
                          &                                      & High                                                                              & 71.92\%                      & 68.86\%                       & 38.09\%                       & 47.42\%                       \\ \hline
\multirow{4}{*}{\de}       & \multirow{2}{*}{from $1e3$ to $1e4$} & Low                                                                               & 11.70\%                      & 12.90\%                       & 16.90\%                       & 16.04\%                       \\
                          &                                      & High                                                                              & 20.84\%                      & 16.83\%                       & 13.58\%                       & 14.06\%                       \\
                          & \multirow{2}{*}{from $1e3$ to $1e5$} & Low                                                                               & 18.61\%                      & 27.02\%                       & 24.40\%                       & 29.66\%                       \\
                          &                                      & High                                                                              & 28.35\%                      & 28.20\%                       & 23.86\%                       & 24.09\%                       \\ \hline
\multirow{4}{*}{\eaeig}   & \multirow{2}{*}{from $1e3$ to $1e4$} & Low                                                                               & 5.90\%                       & 10.51\%                       & 20.12\%                       & -9.40\%                       \\
                          &                                      & High                                                                              & 16.69\%                      & 7.23\%                        & 3.95\%                        & 14.96\%                       \\
                          & \multirow{2}{*}{from $1e3$ to $1e5$} & Low                                                                               & 15.65\%                      & 15.75\%                       & 31.33\%                       & 23.05\%                       \\
                          &                                      & High                                                                              & 22.37\%                      & 7.04\%                        & 19.57\%                       & 9.44\%                        \\ \hline
\multirow{4}{*}{\ebocmar} & \multirow{2}{*}{from $1e3$ to $1e4$} & Low                                                                               & 6.66\%                       & -4.57\%                       & 3.92\%                        & -1.12\%                       \\
                          &                                      & High                                                                              & -11.40\%                     & 6.18\%                        & 6.80\%                        & 13.76\%                       \\
                          & \multirow{2}{*}{from $1e3$ to $1e5$} & Low                                                                               & 16.69\%                      & -6.32\%                       & 4.79\%                        & 4.89\%                        \\
                          &                                      & High                                                                              & -0.13\%                      & 12.16\%                       & 21.77\%                       & 10.62\%                         \\ \hline
\multirow{4}{*}{\elshade}    & \multirow{2}{*}{from $1e3$ to $1e4$} & Low                                                                               & 9.31\%                       & 8.45\%                        & 14.99\%                       & 14.79\%                       \\
                            &                                      & High                                                                              & 10.66\%                      & 8.67\%                        & 17.25\%                       & 6.59\%                        \\
                            & \multirow{2}{*}{from $1e3$ to $1e5$} & Low                                                                               & 9.31\%                       & 3.14\%                        & 11.09\%                       & 15.01\%                       \\
                            &                                      & High                                                                              & 12.57\%                      & 7.25\%                        & 14.24\%                       & 7.91\%                        \\ \hline
\multirow{4}{*}{\dtcg} & \multirow{2}{*}{from $1e3$ to $1e4$} & Low                                                                               & 29.30\%                      & 5.17\%                        & 0.83\%                        & 29.96\%                       \\
                            &                                      & High                                                                              & 3.08\%                       & 9.08\%                        & 27.79\%                       & 37.40\%                       \\
                            & \multirow{2}{*}{from $1e3$ to $1e5$} & Low                                                                               & 29.59\%                      & 9.84\%                        & 3.47\%                        & 46.82\%                       \\
                            &                                      & High                                                                              & 7.70\%                       & 17.43\%                       & 38.22\%                       & 43.12\%                       \\ \hline
\multirow{4}{*}{\lshade}     & \multirow{2}{*}{from $1e3$ to $1e4$} & Low                                                                               & 4.51\%                       & -0.90\%                       & 2.86\%                        & 2.54\%                        \\
                            &                                      & High                                                                              & 15.35\%                      & -1.67\%                       & 1.47\%                        & 2.34\%                        \\
                            & \multirow{2}{*}{from $1e3$ to $1e5$} & Low                                                                               & 4.93\%                       & 3.16\%                        & 3.03\%                        & 3.10\%                        \\
                            &                                      & High                                                                              & 10.47\%                      & 2.01\%                        & 10.99\%                       & 5.02\%                        \\ \hline
\multirow{4}{*}{\nm}         & \multirow{2}{*}{from $1e3$ to $1e4$} & Low                                                                               & 62.50\%                      & 60.88\%                       & 57.20\%                       & 50.24\%                       \\
                            &                                      & High                                                                              & 64.14\%                      & 57.66\%                       & 51.19\%                       & 52.39\%                       \\
                            & \multirow{2}{*}{from $1e3$ to $1e5$} & Low                                                                               & 73.68\%                      & 77.24\%                       & 81.73\%                       & 78.80\%                       \\
                            &                                      & High                                                                              & 76.42\%                      & 77.87\%                       & 70.78\%                       & 69.07\%                       \\ \hline
\multirow{4}{*}{\pso}        & \multirow{2}{*}{from $1e3$ to $1e4$} & Low                                                                               & 9.72\%                       & 1.97\%                        & 3.28\%                        & 15.89\%                       \\
                            &                                      & High                                                                              & 9.30\%                       & 10.37\%                       & 15.59\%                       & 11.30\%                       \\
                            & \multirow{2}{*}{from $1e3$ to $1e5$} & Low                                                                               & 3.94\%                       & -0.85\%                       & 14.38\%                       & 14.60\%                       \\
                            &                                      & High                                                                              & 14.01\%                      & 11.14\%                       & 19.73\%                       & 16.08\%                       \\ \hline
\multirow{4}{*}{\spso}       & \multirow{2}{*}{from $1e3$ to $1e4$} & Low                                                                               & 4.99\%                       & 2.16\%                        & 0.59\%                        & 0.30\%                        \\
                            &                                      & High                                                                              & 4.58\%                       & 1.57\%                        & 0.54\%                        & 0.41\%                        \\
                            & \multirow{2}{*}{from $1e3$ to $1e5$} & Low                                                                               & -2.40\%                      & 17.60\%                       & -4.71\%                       & -18.24\%                      \\
                            &                                      & High                                                                              & 0.93\%                       & -0.95\%                       & -1.29\%                       & -5.85\%                  \\ \hline    

\end{tabular}
\end{table}

When we look at the improvements in increasing $B$ for the individual methods in Table~\ref{perdev} we find additional insights. For instance \nm{} has a remarkable average improvement of 75.6\% when going from $B=1e3$ to $B=1e5$ on $DV=5$. Analyzing the convergence plots of \nm{} reveals that it initially finds non-optimal paths of high-costs in $1e3$, oftentimes crossing multiple threat areas. On the contrary to \nm, \agskimode{} starts with paths that are closer to the best-found ones, see Table~\ref{tab-pe3}. 
%Therefore, the enormous improvement of 36.8\% demonstrated by \agskimode{} \ref{perdev}, further validates its usefulness and promote it as a top choice for $1e5$ \ref{fig:conv}.
Another thing to notice is that \agskimode{} exhibits a higher average improvement in high-density problems, 47\%, compared to its predecessor method \agsk, which improves only a third as much 15.6\%. This is an indicator that \agskimode{} can search faster in high-density (and more complicated) problems.
\birmin{} as a deterministic method continues to improve, at the cost of longer computation time. Particularly, this method had to be truncated at the three-hour limit, which means that the result we see in Table \ref{perdev} is actually not for $1e5$ but rather for $0.91e5, 0.45e5, 0.28e5, 0.28e5$ as an average of the truncation of $DV={5,10,15,20}$ respectively, which is one of the reasons for its relatively poor performance in $B=1e5$ (see also Fig.~\ref{fig:res1e5}). 
%(Jakub, if you are reading the text PDF, please refer to the code commented below)
% i checked this by loading the tables we have and then run this code:

%  for i=1:56
% temp2 = Results_BIRMIN{1, i}.history;
% temp1(i) = temp2(end,2)
% end
% mmean=mean(temp1)
% per = mean(temp1)/dv*3*e5

Another surprising result can be observed in the results for \spso, where the method shows a slight improvement over $1e4$, followed by negative improvements for $1e5$, suggesting that a larger computational budget is ineffective for this method. This behaviour is not, however, present in \pso{} even though we pass spherical coordinate as inputs as well, meaning that the authors of \spso{} had some other unannounced tweaks imposed to improve PSO with faster computation,  as evidenced by the superiority of \spso{} over \pso{} in Fig. \ref{fig:res1e3} and Fig. \ref{fig:res1e4},  with the situation turning around for $B = 1e5$ in Fig. \ref{fig:res1e5}.

%% file: Conclusion.tex
\section{Conclusion}\label{sec5}

In this paper, we investigated the performance of global optimization methods on the UAV path planning problem. To investigate the capabilities of different types of global optimization methods on these UAV problems, we devised a problem generator and picked 56 instances for numerical comparison. For these instances, we computed the ELA features and compared them with ELA features from three established benchmark suits. The results showed that the UAV problems differ significantly from problems in these suits. The UAV problems might serve as a valuable part of real-world problem-based benchmark suits, which are highly desirable \cite{kudela2022critical}. It also would be interesting to measure the representativeness of the different established benchmark suits with respect to the proposed UAV problems \cite{chen2024representativeness}.

We selected twelve representative global optimization methods for the numerical comparisons, which were conducted in varying dimensions and computational budgets. Overall, the best-performing methods (in terms of the three studied criteria: mean relative error, number of wins, and Friedman ranks) were \eaeig, \agskimode, and \elshade, which are all well-performing methods from recent CEC competitions and large-scale benchmarking studies \cite{stripinis2024benchmarking}.

We have also investigated the variable dimension characteristic, which was enabled by the nature of the UAV problems. Although most of the best-found solutions were found in the setting with the lowest dimension ($DV = 5$), there were several ones in the higher dimensions as well (for all computational budgets). 

There is still much further work that could be done. The benchmark problems should be ported into more languages (e.g., Python or C++) and ideally ported into the established benchmarking platforms, such as the IOHprofiler \cite{doerr2018iohprofiler}. Another improvement that could be investigated is an extended computation of the objective function (with finer discretization that could account for collisions with terrain). This computation could also be GPU-accelerated, which would make it faster (and allow for larger computational budgets). We also plan to search for other application-based problems that could serve as benchmarks for variable dimension problems, possibly construction a real-world problem-based suit for benchmarking. Lastly, benchmarking ``proper'' variable dimension global optimization methods (even alongside their fixed dimension counterparts) will be the focus of our future work.

%% file: Acknowledgement.tex
\section*{Acknowledgement}

This work was supported by the project GACR No. 24-12474S ``Benchmarking derivative-free global optimization methods'' and by the project IGA BUT No. FSI-S-23-8394 ``Artificial intelligence methods in engineering tasks''.